%% file: main.tex
\documentclass[10pt,twocolumn,letterpaper]{article}

\usepackage[pagenumbers]{cvpr} %

\definecolor{cvprblue}{rgb}{0.21,0.49,0.74}
\usepackage{float}
\usepackage{graphicx}
\usepackage{amsmath}
\usepackage{amsfonts}
\usepackage{bm}
\usepackage{amssymb}
\usepackage{booktabs}
\usepackage{xcolor}
\usepackage{pifont}
\usepackage{colortbl}
\newcommand{\cmark}{\textcolor{green!80!black}{\ding{51}}}
\newcommand{\xmark}{\textcolor{red}{\ding{55}}}
\usepackage{balance}
\usepackage{wrapfig}

\newcommand{\acro}{Diffusion Self-Distillation\xspace}

\newcommand{\pf}[1]{\cellcolor{red!30}#1}
\newcommand{\ps}[1]{\cellcolor{orange!30}#1}
\newcommand{\pt}[1]{\cellcolor{yellow!30}#1}

\usepackage[pagebackref,breaklinks,colorlinks,allcolors=cvprblue]{hyperref}

\def\paperID{136} %
\def\confName{CVPR}
\def\confYear{2025}

\title{Diffusion Self-Distillation for Zero-Shot Customized Image Generation}

\author{
Shengqu Cai
\quad Eric Ryan Chan
\quad Yunzhi Zhang\\
\quad Leonidas Guibas
\quad Jiajun Wu
\quad Gordon Wetzstein
\\
Stanford University
}

\input{preamble}
\begin{document}
\input{fig/teaser}
\maketitle
\input{sec/0_abstract}    
\input{sec/1_intro}

\input{sec/2_related_work_short}
\input{sec/3_method}
\input{sec/4_result}
\input{sec/5_conclusion}

{
    \clearpage
    \balance
    \small
    \bibliographystyle{ieeenat_fullname}
    \bibliography{main}
}

\input{sec/X_suppl}

\end{document}

%% file: preamble.tex
\usepackage{microtype}  %
\usepackage{flushend}
\usepackage{overpic}
\usepackage{enumitem} %
\usepackage{overpic} %
\usepackage{color}

\definecolor{turquoise}{cmyk}{0.65,0,0.1,0.3}
\definecolor{purple}{rgb}{0.65,0,0.65}
\definecolor{dark_green}{rgb}{0, 0.5, 0}
\definecolor{orange}{rgb}{0.8, 0.6, 0.2}
\definecolor{red}{rgb}{0.8, 0.2, 0.2}
\definecolor{darkred}{rgb}{0.6, 0.1, 0.05}
\definecolor{blueish}{rgb}{0.0, 0.3, .6}
\definecolor{light_gray}{rgb}{0.7, 0.7, .7}
\definecolor{pink}{rgb}{1, 0, 1}
\definecolor{greyblue}{rgb}{0.25, 0.25, 1}

\usepackage{blindtext}

\renewcommand{\paragraph}[1]{\vspace{1em}\noindent\textbf{#1}.}

%% file: fig/teaser.tex
\twocolumn[{%
  \renewcommand\twocolumn[1][]{#1}%
\maketitle
\begin{center}
  \newcommand{\teaserwidth}{0.99\textwidth}
  \vspace{-5pt}
  \centerline{
    \includegraphics[width=\teaserwidth]{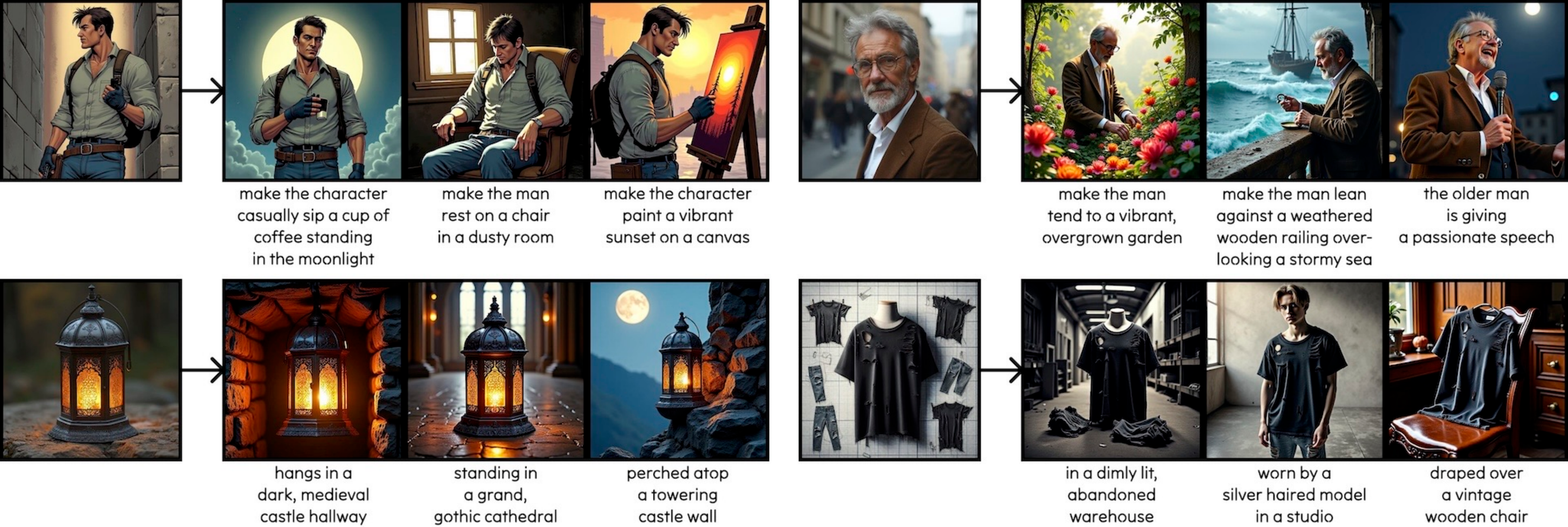}
  }
  \captionsetup{type=figure}
  \vspace{-5pt}
  \captionof{figure}{
    Given an input image, \acro is a novel diffusion-based approach that generates diverse images that maintain the input's identity across various contexts. Unlike prior approaches that require fine-tuning or are limited to specific domains, \acro offers instant customization without any additional inference-stage training, enabling precise control and editability in text-to-image diffusion models. This ability makes \acro a valuable tool for general AI content creation.
  }
  \label{fig:teaser}
 \end{center}%
}]

%% file: sec/0_abstract.tex
\begin{abstract}

\vspace{-5pt}
Text-to-image diffusion models produce impressive results but are frustrating tools for artists who desire fine-grained control. For example, a common use case is to create images of a specific concept in novel contexts, i.e., ``identity-preserving generation''. This setting, along with many other tasks (e.g., relighting), is a natural fit for image+text-conditional generative models. However, there is insufficient high-quality paired data to train such a model directly. We propose \acro, a method for using a pre-trained text-to-image model to generate its own dataset for text-conditioned image-to-image tasks. We first leverage a text-to-image diffusion model's in-context generation ability to create grids of images and curate a large paired dataset with the help of a vision-language model. We then fine-tune the text-to-image model into a text+image-to-image model using the curated paired dataset. We demonstrate that \acro outperforms existing zero-shot methods and is competitive with per-instance tuning techniques on a wide range of identity-preserving generation tasks, without requiring test-time optimization.
Project page: \href{https://primecai.github.io/dsd}{primecai.github.io{\slash}dsd}.
\vspace{-15pt}

\end{abstract}

%% file: sec/1_intro.tex
\section{Introduction}
\label{sec:intro}
In recent years, text-to-image diffusion models~\cite{nichol2021glide,ramesh2022dalle2,saharia2022imagen,rombach2022latentdiffusion} have set new standards in image synthesis, generating high-quality and diverse images from textual prompts.
However, while their ability to generate images from text is impressive, these models often fall short in offering precise control, editability, and consistency—key features that are crucial for real-world applications.
Text input alone can be insufficient to convey specific details, leading to variations that may not fully align with the user’s intent, especially in scenarios that require faithful adaptation of a character or asset’s identity across different contexts.

\input{fig/overview}
Maintaining the instance's identity is challenging, however. We distinguish \textit{structure-preserving} edits, in which the target and source image share the general layout, but may differ in style, texture, or other local features, and \textit{identity-preserving} edits, where assets are recognizably the same across target and source images despite potentially large-scale changes in image structure (Fig.~\ref{fig:edit_vs_customize}). The latter task is a superset of the former and requires the model to have a significantly more profound understanding of the input image and concepts to extract and customize the desired identity. For example, image editing~\cite{brooks2022instructpix2pix, zhang2023controlnet, meng2022sdedit}, such as local content editing, re-lighting, and semantic image synthesis, etc. are all \textit{structure-preserving} and \textit{identity-preserving} edits, but novel-view synthesis and character-consistent generation under pose variations, are \textit{identity-preserving} but not \textit{structure-preserving}. We aim to address the general case, maintaining identity without constraining structure.

For \textit{structure-preserving} edits, adding layers, as in ControlNet~\cite{zhang2023controlnet}, introduces spatial conditioning controls but is limited to structure guidance and does not address consistent identity adaptation across diverse contexts.
For \textit{identity-preserving} edits, fine-tuning methods such as DreamBooth~\cite{ruiz2022dreambooth} and LoRA~\cite{hu2021lora} can improve consistency using a few reference samples but are time consuming and computationally intensive, requiring training for each reference.
Zero-shot alternatives like IP-Adapter~\cite{ye2023ipadapter} and InstantID~\cite{wang2024instantid} offer faster solutions without the need for retraining but fall short in providing the desired level of consistency and customization; IP-Adapter~\cite{ye2023ipadapter} lacks full customization capabilities, and InstantID~\cite{wang2024instantid} is restricted to facial identity.

In this paper, we propose a novel approach called \acro, designed to address the core challenge of zero-shot instant customization and adaptation of any character or asset in text-to-image diffusion models.
We identify the primary obstacle that hinders prior methods, such as IP-Adapter~\cite{ye2023ipadapter} and InstantID~\cite{wang2024instantid}, from achieving better identity preservation or generalizing beyond facial contexts: the absence of large-scale paired datasets and corresponding supervised identity-preserving training pipelines.
With recent advancements in foundational model capabilities, we are now positioned to exploit these strengths further.
Specifically, we can generate consistent grids of identical characters or assets, opening a new pathway for customization that eliminates the need for pre-existing, handcrafted paired datasets—which are expensive and time consuming to collect.
The ability to generate these consistent grids likely emerged from foundational model training on diverse datasets, including photo albums, mangas, and comics.
Our approach harnesses Vision-Language Models~(VLMs) to automatically curate many generated grids, producing a diverse set of grid images with consistent identity features across various contexts.
This curated synthetic dataset then serves as the foundation for fine-tuning and adapting any identity, transforming the task of zero-shot customized image generation from unsupervised to supervised.
\acro offers transformative potential for applications like consistent character generation, camera control, relighting, and asset customization in fields such as comics and digital art.
This flexibility allows artists to rapidly iterate and adapt their work, reducing effort and enhancing creative freedom, making \acro a valuable tool for AI-generated content.

We summarize our contributions as follows:
\begin{itemize}
    \item We propose \acro, a \emph{zero-shot} identity-preserving customized image generation model that scales to any instance under any context, with performances on par with inference-stage tuning methods;
    \item We provide a self-distillation pipeline to obtain identity-preserving data pairs purely from pretrained text-to-image diffusion models, LLMs, and VLMs, without any human effort involved in the entire data creation wheel;
    \item We correspondingly design a unified architecture for image-to-image translation tasks involving \emph{both} identity- and structure-preserving edits, including personalization, relighting, depth controls, and instruction following.
\end{itemize}
\input{fig/edit_vs_customize}

%% file: fig/overview.tex
\begin{figure*}[ht]
\begin{center}
\centering
\includegraphics[width=0.99\linewidth]{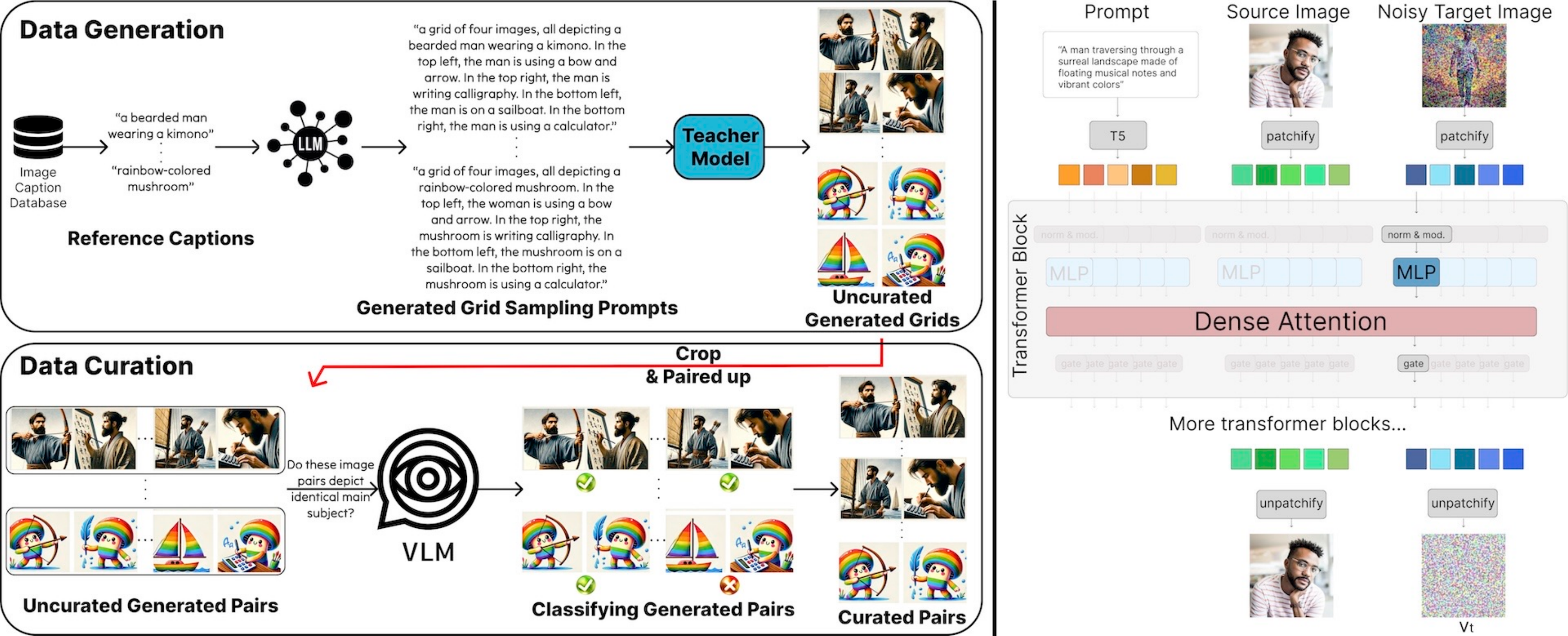}
\end{center}
\vspace{-15pt}
\caption{
\textbf{Overview of our pipeline.}
\textit{Left:}
the top shows our vanilla paired data generation wheel~(Sec.~\ref{sec:data_generation}).
We first sample reference image captions from the LAION~\cite{schuhmann2022laion} dataset.
These reference captions are parsed through an LLM to be translated into identity-preserved grid generation prompts~(Sec.~\ref{sec:prompt_generation}).
We feed these enhanced prompts to a pretrained text-to-image diffusion model to sample potentially identity-preserved grids of images, which are then cropped and composed into vanilla image pairs~(Sec.~\ref{sec:vanilla_data_generation}).
On the bottom, we show our data curation pipeline~(Sec.~\ref{sec:data_curation}), where the vanilla image paired are fed into a VLM to classify whether they depict identical main subjects. This process mimics a human annotation/curation process while being fully automatic; we use the curated data as our final training data.
\textit{Right:}
we extend the diffusion transformer model into an image-conditioned framework by treating the input image as the first frame of a two-frame sequence.
The model generates both frames simultaneously—the first reconstructs the input, while the second is the edited output—allowing effective information exchange between the conditioning image and the desired output.
}
\vspace{-15pt}
\label{fig:overview}
\end{figure*}

%% file: fig/edit_vs_customize.tex
\begin{figure}[t]
\begin{center}
\centering
\includegraphics[width=0.99\linewidth]{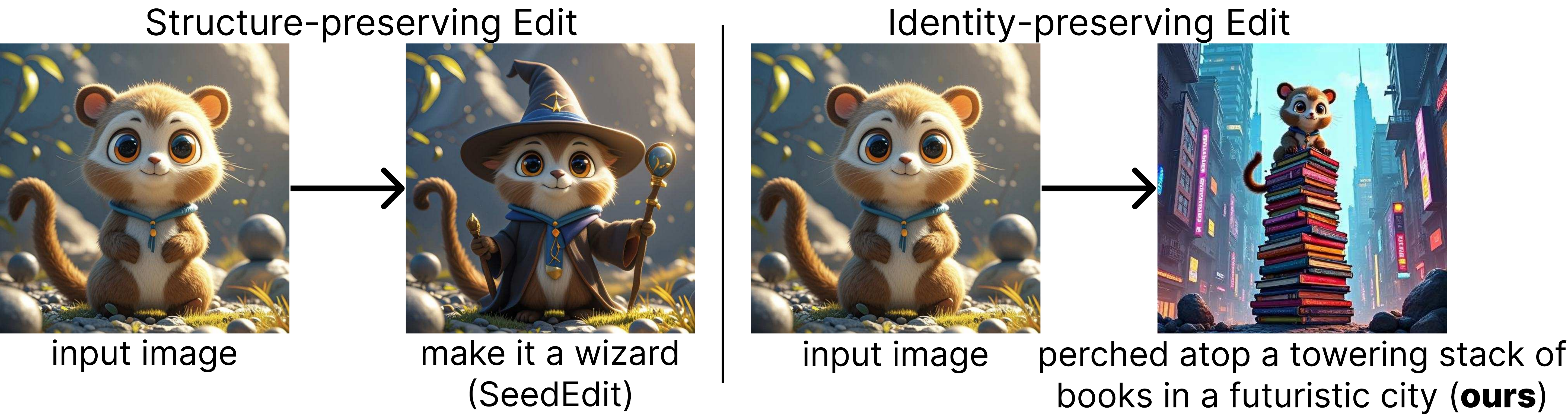}
\end{center}
\vspace{-20pt}
\caption{
\textbf{Difference between \textit{structure-preserving} and \textit{identity-preserving} edits. }
In \textit{structure-preserving} editing, the main structures of the image are preserved, and only local edits or stylizations are performed.
In \textit{identity-preserving} editing, the global structure of the image may change radically.
}
\vspace{-15pt}
\label{fig:edit_vs_customize}
\end{figure}

%% file: sec/2_related_work_short.tex
\section{Related work}
\label{sec:related_work}
Recent advancements in diffusion models have underscored the need for enhanced control and customization in image-generation tasks. Various methods have been proposed to address these challenges through additional conditioning mechanisms, personalization, and rapid adaptation~\citep{po2023state}.

\paragraph{Control Mechanisms in Diffusion Models} To move beyond purely text-based controls, approaches like ControlNet~\cite{zhang2023controlnet} introduce spatial conditioning via inputs such as sketches, depth maps, and segmentation masks, enabling fine-grained structure control. ControlNet++~\cite{li2024controlnetplusplus} refines this by enhancing the integration of spatial inputs for more nuanced control. Uni-ControlNet~\cite{zhao2023unicontrolnet} unifies various control types within a single framework, standardizing the handling of diverse signals. T2I-Adapter~\cite{mou2023t2iadapter} employs lightweight adapters to align pretrained models with external control signals without altering the core architecture. While these methods offer increased flexibility, they often focus on structural conditioning types such as depths and lack capabilities for concept extraction or identity preservation.

\paragraph{Personalization and Fine-Tuning} Techniques like DreamBooth~\cite{ruiz2022dreambooth} and LoRA~\cite{hu2021lora} enhance the consistency and relevance of generated images by fine-tuning models with small sets of reference images. DreamBooth~\cite{ruiz2022dreambooth} personalizes models to maintain a subject's identity across different contexts, while LoRA~\cite{hu2021lora} provides an efficient approach to fine-tuning large models without extensive retraining. However, these methods require multiple images and test-time optimization for each reference, which can be computationally expensive—especially with the exponential growth in model sizes~(12 billion parameters for FLUX).

\paragraph{Zero-Shot and Fast Adaptation} IP-Adapter~\cite{ye2023ipadapter} incorporates image prompts into diffusion models using image embeddings, allowing for generations that align closely with reference visuals. InstantID~\cite{wang2024instantid} ensures zero-shot face preservation, maintaining a subject's key features across various contexts. While effective as zero-shot methods without user training, IP-Adapter~\cite{ye2023ipadapter} struggles to adapt specific targets like unique characters or assets, and InstantID~\cite{wang2024instantid} is limited to facial identity preservation. IPAdapter-Instruct~\cite{rowles2024ipadapterinstruct} enhances image-based conditioning with instruct prompts but relies heavily on specific instructions and task-specific pretrained models. Other methods, such as SuTI~\cite{chen2023suti} and GDT~\cite{huang2024gdt}, handcrafted corresponding datasets, which are expensive and challenging to collect and scale. Another work along this line is Subject-Diffusion~\cite{ma2024subjectdiffusion}, which uses segmentation masks to create synthetic data for training but is bounded by achieving only simple attributes and accessories copying and editing within input images.

Existing methods contribute valuable advancements but often target specific domains or require user-stage tuning. \acro{} bridges these gaps by offering a unified, zero-shot approach for consistent customization of characters and assets using minimal input. By leveraging self-distillation assisted by vision-language models, \acro{} provides a comprehensive and adaptable solution for a wide range of creative applications.

%% file: sec/3_method.tex
\section{\acro}
\label{sec:method}
We discover that recent text-to-image generation models offer the surprising ability to generate in-context, consistent image grids~(see Fig.~\ref{fig:overview}, left).
Motivated by this insight, we develop a zero-shot adaptation network that offers fast, diverse, high-quality, and identity-preserving, i.e., consistent image generation conditioned on a reference image. 
For this purpose, we first generate and curate sets of images that exhibit the desired consistency using pretrained text-to-image diffusion models, large language models~(LLMs), and vision-language models~(VLMs)~(Sec.~\ref{sec:data_generation}).
Then, we finetune the same pretrained diffusion model with these consistent image sets, employing our newly proposed parallel processing architecture~(Sec.~\ref{sec:architecture}) to create a conditional model.
By this end, \acro finetunes a pretrained text-to-image diffusion model into a zero-shot customized image generator in a supervised manner.
\subsection{Generating a Pairwise Dataset} \label{sec:data_generation}
To create a pairwise dataset for supervised \acro training, we leverage the emerging multi-image generation capabilities of pretrained text-to-image diffusion models to produce potentially consistent vanilla images~(Sec.~\ref{sec:vanilla_data_generation}) created by LLM-generated prompts~(Sec.~\ref{sec:prompt_generation}).
We then use VLMs to curate these vanilla samples, obtaining clean sets of images that share the desired identity consistency~(Sec.~\ref{sec:data_curation}). The data generation and curation pipeline is shown in Fig.~\ref{fig:overview}, left.
\subsubsection{Vanilla Data Generation via Teacher Model} \label{sec:vanilla_data_generation}
To generate sets of images that fulfill the desired identity preservation, we prompt the teacher pretrained text-to-image diffusion model to create images containing multiple panels featuring the same subject with variations in expression, pose, lighting conditions, and more, for training purposes.
Such prompting can be as simple as specifying the desired identity preservation in the output, such as ``\textit{a grid of 4 images representing the same $<object/character/scene/etc.>$}'', ``\textit{an evenly separated 4 panels, depicting identical $<object/character/scene/etc.>$}'', etc.
We additionally specify the expected content in each sub-image/panel.
The full set of prompts is provided in our supplemental material Sec.~\ref{sec:data_pipeline_prompts}.
Our analysis shows that current state-of-the-art text-to-image diffusion models~(e.g., SD3~\cite{esser2024sd3}, DALL·E 3, FLUX) demonstrate this identity-preserving capability, likely emerging from their training data, which includes comics, mangas, photo albums, and video frames.
Such in-context generation ability is crucial to our data generation wheel.
\subsubsection{Prompt Generation via LLMs} \label{sec:prompt_generation}
We rely on an LLM to ``brainstorm'' a large dataset of diverse prompts, from which we derive our image grid dataset.
By defining a prompt structure, we prompt the LLM to produce text prompts that describe image grids.
A challenge we encountered is that when prompted to create large sets of prompts, LLMs tend to produce prompts of low diversity.
For example, we noticed that without additional guidance, GPT-4o has a strong preference for prompts with cars and robots, resulting in highly repetitive outputs.
To address this issue, we utilize the available image captions in the LAION~\cite{schuhmann2022laion} dataset, feeding them into the LLM as content references.
These references from real image captions dramatically improve the diversity of generated prompts.
Optionally, we also use the LLM to filter these reference captions, ensuring they contain a clear target for identity preservation.
We find that this significantly improves the hit rate of generating consistent multi-image outputs.
\subsubsection{Dataset Curation and Caption with VLMs} \label{sec:data_curation}
While the aforementioned data generation scheme provides identity-preserving multi-image samples of decent quality and quantity, these initial ``uncurated'' images tend to be noisy and unsuitable for direct use.
Therefore, we leverage the strong capabilities of VLMs to curate a clean dataset.
We extract pairs of images from the generated samples intended to preserve the identity and ask the VLM whether the two images depict the same object, character, scene, etc.
We find that employing Chain-of-Thought prompting~\cite{wei2022chainofthought} is particularly helpful in this context.
Specifically, we first prompt the VLM to identify the common object, character, or scene present in both images, then have it describe each one in detail, and finally analyze whether they are identical, providing a conclusive response.
This process yields pairs of images that share the same identity.

\subsection{Parallel Processing Architecture} \label{sec:architecture}
We desire a conditional architecture suitable for general image-to-image tasks, including transformations in which structure is preserved, and transformations in which concepts/identities are preserved but image structure is not.
This is a challenging problem because it may necessitate the transfer of fine details without guaranteeing spatial correspondences.
While the ControlNet~\cite{zhang2023controlnet} architecture is excellent at structure-preserving edits, such as depth-to-image or segmentation-map-to-image, it struggles to preserve details under more complex identity-preserving edits, where the source and target images are not pixel-aligned.
On the other hand, IP-Adapter~\cite{ye2023ipadapter} can extract certain concepts, such as styles, from the input image.
Still, it strongly relies on a task-specific image encoder and often fails to preserve more complex concepts and identities.
Drawing inspiration from the success of multi-view and video diffusion models~\cite{guo2023animatediff,cai2023genren,kuang2024cvd, brooks2024sora,blattmann2023svd,yang2024cogvideox,hong2022cogvideo,cai2023diffdreamer,li2023instant3d,xu2023dmv3d,shao2024human4dit,he2022lvdm,shi2023mvdream,hu2023animateanyone}, we propose a simple yet effective method to extend the vanilla diffusion transformer model into an image-conditioned diffusion model.
Specifically, we treat the input image as the first frame of a video and produce a two-frame video as output.
The final loss is computed over the two-frame video, establishing an identity mapping for the first frame and a conditionally editing target for the second frame.
Our architecture design allows generality for generic image-to-image translation tasks, since it enables effective information exchange between the two frames, allowing the model to capture complex semantics and perform sophisticated edits, as shown in Fig.~\ref{fig:overview}, right.

%% file: sec/4_result.tex
\input{fig/qualitative_comparison}
\input{fig/qualitative_result}
\section{Experiments}
\label{sec:result}
\paragraph{Implementation details}
We use FLUX1.0 DEV as both our teacher and student models, achieving self-distillation. For prompt generation, we use GPT-4o; for dataset curation and captioning, we use Gemini-1.5. We train all models on 8 NVIDIA H100 80GB GPUs with an effective batch size of 160 for 100k iterations, using AdamW optimizer~\cite{loshchilov2018adamw} with a learning rate $10^{-4}$. Our parallel processing architecture uses LoRAs with rank 512 on the base model.

\paragraph{Datasets}
Our final training dataset contains $\sim$ 400k subject-consistent image pairs generated from our teacher model, FLUX1.0 DEV. Generating and curating the dataset is fully automated and requires no human effort, so its size could be further scaled.
We use the publicly available DreamBench++~\cite{peng2024dreambenchplus} dataset and follow their protocols for evaluation. DreamBench++~\cite{peng2024dreambenchplus} is a comprehensive and diverse dataset for evaluating personalized image generation, consisting of 150 high-quality images and 1,350 prompts—significantly more than previous benchmarks like DreamBench~\cite{ruiz2022dreambooth}. The dataset covers various categories such as animals, humans, objects, etc., including photorealistic and non-photorealistic images, with prompts designed to span different difficulty levels~(simple/imaginative). In contrast, prompts are generated using GPT-4o and refined by human annotators to ensure diversity and ethical compliance.

\paragraph{Baselines}
We follow the setups in DreamBench++~\cite{peng2024dreambenchplus} and compare our model with two classes of baselines: inference-stage tuning models and zero-shot models.
For inference-stage models, we compare against Textual Inversion~\cite{gal2023textualinversion}, DreamBooth~\cite{ruiz2022dreambooth} and its LoRA~\cite{hu2021lora} version.
For zero-shot models, we compare with BLIP-Diffusion~\cite{li2023blipdiffusion}, Emu2~\cite{sun2023emu2}, IP-Adapter~\cite{ye2023ipadapter}, IP-Adapter+~\cite{ye2023ipadapter}.

\paragraph{Evaluation metrics}
The evaluation protocol of prior works~\cite{ye2023ipadapter,rowles2024ipadapterinstruct,ruiz2022dreambooth,chen2023suti} typically involves comparing the CLIP~\cite{radford2021clip} and DINO~\cite{caron2021dinov1} feature similarities.
However, we note that the metrics mentioned above capture only global semantic similarity, are extremely noisy, and are biased towards ``copy-pasting'' the input image.
This is especially troublesome when the input image or the prompt is complex.
We refer to DreamBench++~\cite{peng2024dreambenchplus} for a detailed analysis of their limitations.
Therefore, we follow the metrics designed in DreamBench++~\cite{peng2024dreambenchplus} and report GPT-4o scores on the more diverse DreamBench++~\cite{peng2024dreambenchplus} benchmark for both concept preservation~(CP) with different categories of subjects and prompt following~(PF) with photorealistic~(Real.) and Imaginative~(Imag.) prompts, then use their product as a final evaluation score.
This evaluation protocol emulates a human user study using VLMs.
We additionally slightly modify the GPT evaluation prompts so that penalization can be applied if the generated contents show no internal understanding and creative output but instead naively copy over components from the reference image.
The modified metrics are named ``de-biased concept preservation~(Debiased CP)'' and ``de-biased prompt following~(Debiased PF)''.
The full set of GPT evaluation prompts will be provided in our supplementary Sec.~\ref{sec:gpt_evaluation_prompts}.

\input{tab/quantitative_result_gpt}
\paragraph{Qualitative results}
Fig.~\ref{fig:qualitative_comparison} presents our qualitative comparison results, demonstrating that our model significantly outperforms all baselines in subject adaptation and concept consistency while exhibiting excellent prompt alignment and diversity in the outputs.
Textual Inversion~\cite{gal2023textualinversion}, as an early concept extraction method, captures only vague semantics from the input image, making it unsuitable for zero-shot customization tasks that require precise subject adaptation.
DreamBooth~\cite{ruiz2022dreambooth} and DreamBooth-LoRA~\cite{ruiz2022dreambooth,hu2021lora} face challenges in maintaining consistency, primarily because they perform better with multiple input images. This dependency limits their effectiveness when only a single reference image is available.
In contrast, our method achieves robust results even with just one input image, highlighting its efficiency and practicality.
BLIP-Diffusion~\cite{li2023blipdiffusion}, operating as a self-supervised representation learning framework, can extract concepts from the input in a zero-shot manner but is confined to capturing overall semantic concepts without the ability to customize specific subjects.
Similarly, Emu2~\cite{sun2023emu2}, a multi-modal foundation model, excels at extracting semantic concepts but lacks mechanisms for specific subject customization, limiting its utility in personalized image generation.
IP-Adapter~\cite{ye2023ipadapter} and IP-Adapter+~\cite{ye2023ipadapter} employ self-supervised learning schemes aimed at reconstructing the input from encoded signals. While effective in extracting global concepts, they suffer from a pronounced ``copy-paste'' effect, where the generated images closely resemble the input without meaningful transformation. Notably, IP-Adapter+~\cite{ye2023ipadapter}, which utilizes a stronger input image encoder, exacerbates this issue, leading to less diversity and adaptability in the outputs.
In contrast, our approach effectively preserves the subject's core identity while enabling diverse and contextually appropriate transformations.
As illustrated in Fig.~\ref{fig:qualitative_result}, our \acro demonstrates remarkable versatility, adeptly handling various customization targets across different targets~(characters, objects, etc.) and styles~(photorealistic, animated, etc.).
Moreover, \acro generalizes well to a wide range of prompts, including instructions similar to InstructPix2Pix~\cite{brooks2022instructpix2pix}, underscoring its robustness and adaptability in diverse customization tasks.

\paragraph{Quantitative results}
Quantitative comparison with the baselines are shown in Tab.~\ref{tab:quantitative_result_gpt}, where we report GPT evaluation following DreamBench++~\cite{peng2024dreambenchplus}.
Such an evaluation protocol is similar to human score but uses automatic multimodal LLMs.
Our method achieves the best overall performances accommodating both concept preservation and prompt following, while only being inferior to IP-Adapter+~\cite{ye2023ipadapter} for the former~(mainly because of the ``copy-paste'' effect again), and the per-instance tuning DreamBooth-LoRA~\cite{ruiz2022dreambooth,hu2021lora} for the latter.
We note that the concept preservation evaluation of DreamBench++~\cite{peng2024dreambenchplus} is still biased towards favoring a ``copy-paste'' effect, especially on more challenging and diverse prompts.
For instance, the outstanding concept preservation performances of the IP-Adapter family~\cite{ye2023ipadapter} are primarily because of their strong ``copy-paste'' effect, which copies over the input image without considering relevant essential changes in the prompts.
This can also be partly observed by their underperforming prompt following scores, which means they are biased towards the reference input and do not accommodate the input prompt.
Therefore, we also present our ``de-biased'' version of GPT scores, which are as simple as telling GPT to penalize if the generated image resembles a direct copy of the reference image.
We observe that the advantages of IP-Adaper+~\cite{ye2023ipadapter} no longer hold.
Overall, \acro is the best-performing model.

\paragraph{Ablation studies}
\input{fig/ablation_study}
\textbf{(1)} \textit{Data curation:}
During dataset generation, we first synthesize grids using a frozen pre-trained FLUX model and then filter the images via VLM curation. Why not fine-tune the FLUX model on image grids to improve the hit rate? To study this, we fit a LoRA~\cite{hu2021lora} using $>7000$ consistent grids (Fig.~\ref{fig:ablation_study}, left).
Though a more significant proportion of samples are consistent grids, we find that the teacher model loses diversity in its output. Therefore, we choose to rely entirely on VLMs to help us curate from large numbers of diverse but potentially noisy grids.
\textbf{(2)} \textit{Parallel processing architecture:}
We compare the parallel processing architecture to three alternative image-to-image architectures: 1) concatenating the source image to the noise image (``concatenation''); 2) a ControlNet~\cite{zhang2023controlnet}-based design, and 3) an IP-Adapter~\cite{ye2023ipadapter}-based design. We train each architecture using the same data as our parallel processing model (Fig.~\ref{fig:ablation_study}, middle).
For ControlNet~\cite{zhang2023controlnet}, we draw the same conclusion as prior work~\cite{hu2023animateanyone}, in that it works best for structure-aligned edits, but generally struggles to preserve details when the source image and target image differ in camera pose.
IP-Adapter~\cite{ye2023ipadapter} struggles to effectively transfer details and styles from the source image due to the limited capacity of its image encoder.
\textbf{(3)} \textit{Other image-to-image tasks:}
Although not ``self-distillation'', since it requires an externally-sourced paired dataset~(generated with Depth Anything~\cite{yang2024depthanything}), we additionally train our architecture on depth-to-image to demonstrate its utility for more general image-to-image tasks (Fig.~\ref{fig:ablation_study}, right).

\paragraph{User study}
\input{tab/user_study}
To evaluate the fidelity and prompt consistency of our generated images, we conducted a user study on a random subset of the DreamBench++~\cite{peng2024dreambenchplus} test cases, selecting 20 samples.
A total of 25 female and 29 male annotators, aged from 22 to 78 (average 34), independently scored each image from 1 to 5 based on three criteria: (1) concept preservation—the consistency with the reference image, (2) prompt alignment—the consistency with the given prompt, and (3) creativity—the level of internal understanding and transformation. The average scores are presented in Tab.~\ref{tab:user_study}.
Our human annotations closely align with the GPT evaluation, demonstrating that our \acro is slightly behind IP-Adapter+\cite{ye2023ipadapter} in concept preservation and the inference-stage tuning method DreamBooth-LoRA~\cite{ruiz2022dreambooth,hu2021lora} in prompt alignment.
Notably, our model achieved the highest creativity score, while IP-Adapter+~\cite{ye2023ipadapter} scored lower in this metric due to its “copy-paste” effect. These results further confirm that our \acro offers the most balanced and superior overall performance.

%% file: fig/qualitative_comparison.tex
\begin{figure*}[t]
\begin{center}
\centering
\includegraphics[width=0.99\linewidth]{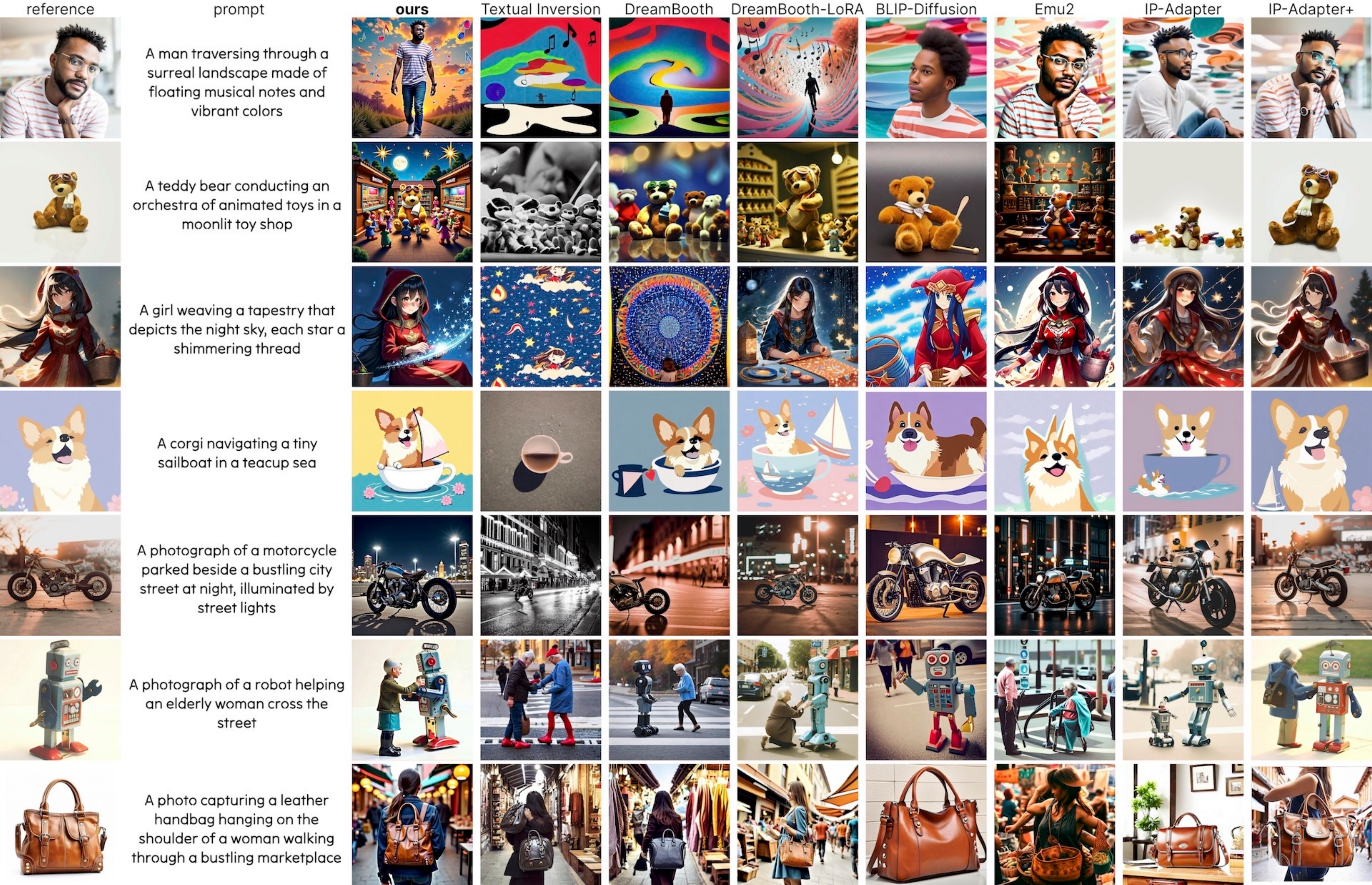}
\end{center}
\vspace{-15pt}
\caption{
\textbf{Qualitative comparison.}
Overall, our method achieves high subject identity preservation and prompt-aligned diversity while not suffering from a ``copy-paste'' effect, such as the results of IP-Adapter+~\cite{ye2023ipadapter}. This is largely thanks to our supervised training pipeline, which alleviates the base model's in-context generation ability.
}
\vspace{-15pt}
\label{fig:qualitative_comparison}
\end{figure*}

%% file: fig/qualitative_result.tex
\begin{figure*}[t]
\begin{center}
\centering
\includegraphics[width=0.99\linewidth]{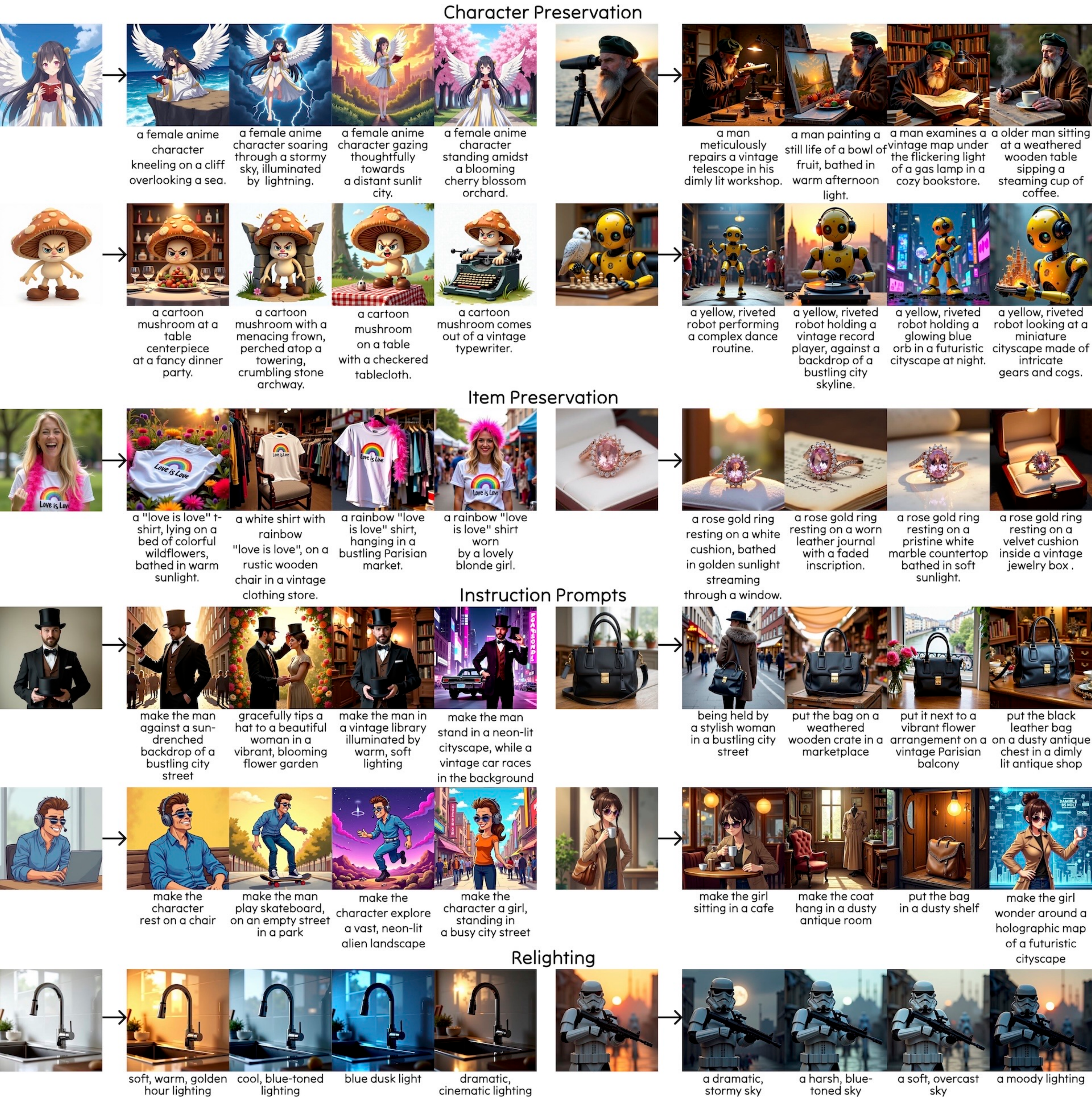}
\end{center}
\vspace{-15pt}
\caption{
\textbf{Qualitative result. }
Our \acro is capable of various customization targets across different tasks and styles, for instance, characters or objects, photorealistic or animated.
\acro can also take instruction types of prompts as input, similar to InstructPix2Pix~\cite{brooks2022instructpix2pix}.
Further, our model exhibits relighting capabilities without significantly altering the scene's content.
}
\vspace{-15pt}
\label{fig:qualitative_result}
\end{figure*}

%% file: tab/quantitative_result_gpt.tex
\begin{table*}[ht]
\centering
\scriptsize
\setlength{\tabcolsep}{1.2pt}
\begin{tabular}{l c c c c c | c c c | c | c c c c | c c c | c}
\toprule
&  & \multicolumn{4}{c|}{Concept Preservation} & \multicolumn{3}{c|}{Prompt Following} & & \multicolumn{4}{c|}{Debiased Concept Preservation} & \multicolumn{3}{c|}{Debiased Prompt Following} & Debiased \\
Method & Z-S? & Animal$\uparrow$ & Human$\uparrow$ & Object$\uparrow$ & Overall$\uparrow$ & Real.$\uparrow$ & Imag.$\uparrow$ & Overall$\uparrow$ & CP$\cdot$PF$\uparrow$ & Animal$\uparrow$ & Human$\uparrow$ & Object$\uparrow$ & Overall$\uparrow$ & Real.$\uparrow$ & Imag.$\uparrow$ & Overall$\uparrow$ & CP$\cdot$PF$\uparrow$
\\
\midrule
Textual Inversion                                     & \xmark &    0.502   &    0.358   &    0.305   &    0.388   &    0.671   &    0.437   &    0.598   &    0.232   &    0.741   &    0.694   &    0.717   &    0.722   &    0.619   &    0.385   &    0.541   & 0.391 \\
DreamBooth                  & \xmark &    0.640   &    0.199   &    0.488   &    0.442   & \ps{0.798} &    0.504   & \pt{0.692} &    0.306   &    0.670   &    0.362   &    0.676   &    0.626   & \pt{0.750} &    0.467   & \pt{0.656} & 0.411 \\
DreamBooth LoRA & \xmark & \ps{0.751} &    0.311   & \pt{0.543} &    0.535   & \pf{0.898} & \pf{0.754} & \pf{0.849} & \ps{0.450} &    0.681   &    0.675   & \pf{0.761} &    0.720   & \pf{0.865} & \pf{0.718} & \pf{0.816} & \ps{0.588} \\
\midrule
BLIP-Diffusion & \cmark &    0.637   &    0.557   &    0.469   &    0.554   &    0.581   &    0.303   &    0.464   &    0.257   & \pt{0.771} & \pt{0.733} & \pt{0.745} & \pt{0.750} &    0.529   &    0.266   &    0.442   & 0.332 \\
Emu2                                                  & \cmark & \pt{0.670} &    0.546   &    0.447   &    0.554   &    0.732   & \pt{0.560} &    0.670   & \pt{0.371} &    0.652   &    0.683   &    0.701   &    0.681   &    0.686   & \pt{0.494} &    0.622   & 0.424\\
IP-Adapter                     & \cmark &    0.667   & \pt{0.558} &    0.504   & \pt{0.576} &    0.743   &    0.446   &    0.607   &    0.350   & \ps{0.790} & \ps{0.764} &    0.743   & \ps{0.766} &    0.695   &    0.377   &    0.589   & \pt{0.451}\\
IP-Adapter+                   & \cmark & \pf{0.900} & \pf{0.845} & \pf{0.759} & \pf{0.834} &    0.502   &    0.279   &    0.388   &    0.324   &    0.481   &    0.473   &    0.530   &    0.504   &    0.442   &    0.229   &    0.371   & 0.187 \\
\textbf{Ours}                                         & \cmark & \underline{0.647} & \ps{\underline{0.567}} & \ps{\underline{0.640}} & \ps{\underline{0.631}}  & \pt{\underline{0.777}} & \ps{\underline{0.625}} & \ps{\underline{0.726}} & \pf{\underline{0.458}} &  \pf{\underline{0.852}}  &  \pf{\underline{0.774}}  & \ps{\underline{0.750}} & \pf{\underline{0.789}}  & \ps{\underline{0.808}} & \ps{\underline{0.681}}  & \ps{\underline{0.757}} & \pf{\underline{0.597}} \\
\bottomrule
\end{tabular}
\vspace{-5pt}
\caption{\textbf{Quantitative result}. 
On the human-aligned GPT score metrics, our method is only inferior to IP-Adapter+~\cite{ye2023ipadapter} for concept preservation~(largely because of IP-Adapter families' ``copy-pasting'' effect) and the tuning-base DreamBooth-LoRA~\cite{ruiz2022dreambooth,hu2021lora} for prompt following, but outperforms every other baseline, achieving the best overall performance considering both concept preservation and prompt following.
We also note that on the de-biased GPT evaluation, which penalizes ``copy-pasting'' the reference image without significant creative interpretation or transformation, the advantages of IP-Adaper+~\cite{ye2023ipadapter} no longer hold.
This can also be partly observed by their bad prompt following scores, meaning they are biased towards the reference input and are not accommodating the input prompt.
The \colorbox{red!40}{\strut first}, \colorbox{orange!50}{\strut second}, and \colorbox{yellow!50}{\strut third} values are highlighted, where \acro{} is the best overall performing model.
}
\vspace{-15pt}
\label{tab:quantitative_result_gpt}  
\end{table*}

%% file: fig/ablation_study.tex
\begin{figure*}[t]
\begin{center}
\centering
\includegraphics[width=0.99\linewidth]{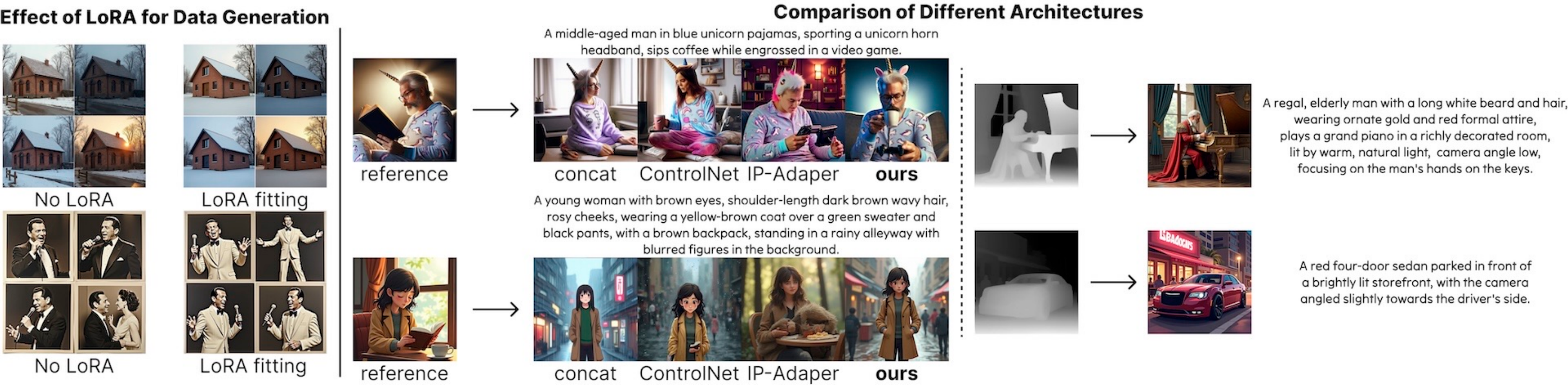}
\end{center}
\vspace{-17pt}
\caption{
\textbf{Ablation study.}
Left: We compare the base model's in-context sampling ability with a consistent grid LoRA-overfitted model. We observe that although applying LoRA to the base model can increase the likelihood of outputs being consistent grids, it may adversely affect output diversity. Therefore, we rely on vision-language models (VLMs) to curate from a large number of diverse but potentially noisy grids.
Right: We compare our architectural design with a vanilla conditional model (by adding a few input channels), ControlNet~\cite{zhang2023controlnet}, and IP-Adapter~\cite{ye2023ipadapter}. Our architecture learns the input concepts and identities significantly better.
We also demonstrate that our architecture can effectively scale to depth-conditioned image generation similar to ControlNet~\cite{zhang2023controlnet}.
}
\vspace{-15pt}
\label{fig:ablation_study}
\end{figure*}

%% file: tab/user_study.tex
\begin{table}[t]
\centering
\resizebox{\linewidth}{!}{ %
\begin{tabular}{l c c c}
\toprule
Method & CP $\uparrow$ & PF $\uparrow$  & Creativity $\uparrow$
\\
\midrule
Textual Inversion~\cite{gal2023textualinversion}      & 1.693 & 1.924 & 2.850 \\
DreamBooth~\cite{ruiz2022dreambooth}                  & 2.329 & \pt{2.883} & \pt{3.597} \\
DreamBooth LoRA~\cite{ruiz2022dreambooth,hu2021lora}  & \pt{2.576} & \pf{3.386} & \ps{4.247} \\
BLIP-Diffusion~\cite{li2023blipdiffusion}             & 1.854 & 2.281 & 0.286 \\
Emu2~\cite{sun2023emu2}                               & 1.843 & 2.096 & 2.965 \\
IP-Adapter~\cite{ye2023ipadapter}                     & 2.274 & 2.307 & 3.481 \\
IP-Adapter+~\cite{ye2023ipadapter}                    & \pf{3.733} & 1.959 & 2.428 \\
\textbf{Ours}               & \ps{\underline{3.661}} & \ps{\underline{3.328}} & \pf{\underline{4.453}} \\
\bottomrule
\end{tabular}
} %
\vspace{-5pt}
\caption{\textbf{User study}. 
``CP'' refers to concept preservation scores and ``PF'' refers to prompt following scores.
The \colorbox{red!40}{\strut first}, \colorbox{orange!50}{\strut second}, and \colorbox{yellow!50}{\strut third} values are highlighted.
Our user study results mostly align with our GPT evaluation, where our \acro{} is the best overall performing model.
}\vspace{-16pt}
\label{tab:user_study}  
\end{table}

%% file: sec/5_conclusion.tex
\vfill
\section{Discussion}
We present \acro, a zero-shot approach designed to achieve identity adaptation across a wide range of contexts using text-to-image diffusion models without any human effort.
Our method effectively transforms zero-shot customized image generation into a supervised task, substantially reducing its difficulty.
Empirical evaluations demonstrate that \acro performs comparably to inference-stage tuning techniques while retaining the efficiency of zero-shot methods.

\paragraph{Limitations and future work}
Our work focuses on identity-preserving edits of characters, objects, and scene relighting.
Future directions could explore additional tasks and use cases.
Integration with ControlNet~\cite{zhang2023controlnet}, for example, could provide fine-grained and independent control of identity and structure.
Additionally, extending our approach from image to video generation is a promising avenue of future work.

\paragraph{Ethics}
We are mindful of the potential misuse, particularly in deepfakes.
We oppose exploiting our work for purposes that infringe upon ethical standards or privacy.

\paragraph{Conclusion}
Our \acro democratizes content creation, enabling identity-preserving, high-quality, and fast customized image generation that adapts seamlessly to evolving foundational models, significantly expanding the creative boundaries of art, design, and digital storytelling.

\paragraph{Acknowledgements}
This project was in part supported by Google, Kaiber AI, 3bodylabs, ONR N00014-23-1-2355, and NSF RI \#2211258. ERC was supported by the Nvidia Graduate Fellowship and the Snap Research Fellowship. YZ was in part supported by the Stanford Interdisciplinary Graduate Fellowship.

%% file: sec/X_suppl.tex
\clearpage
\setcounter{page}{1}
\maketitlesupplementary

\appendix

\section{Data Pipeline Prompts}
\label{sec:data_pipeline_prompts}
In this section, we list out the detailed prompts used in our data generation~(Sec.~\ref{sec:data_generation_prompts}), curation~(Sec.~\ref{sec:data_curation_prompts}) and caption~(Sec.~\ref{sec:image_caption_prompts}) pipelines.
\subsection{Data Generation Prompts}
\label{sec:data_generation_prompts}
To generate grid prompts, we employ GPT-4o as our language model~(LLM) engine
We instruct the LLM to focus on specific aspects during the grid generation process: preserving the identity of the subject, providing detailed content within each grid quadrant, and maintaining appropriate text length.
However, we observed that not all sampled reference captions inherently include a clear instance suitable for identity preservation.
To address this issue, we introduce an initial filtering stage to ensure that each sampled reference caption contains an identity-preserving target.
This filtering enhances the quality and consistency of the generated grids.
\begin{figure}[H]
\small
\begin{center}
{\small
\includegraphics[width=0.99\columnwidth]{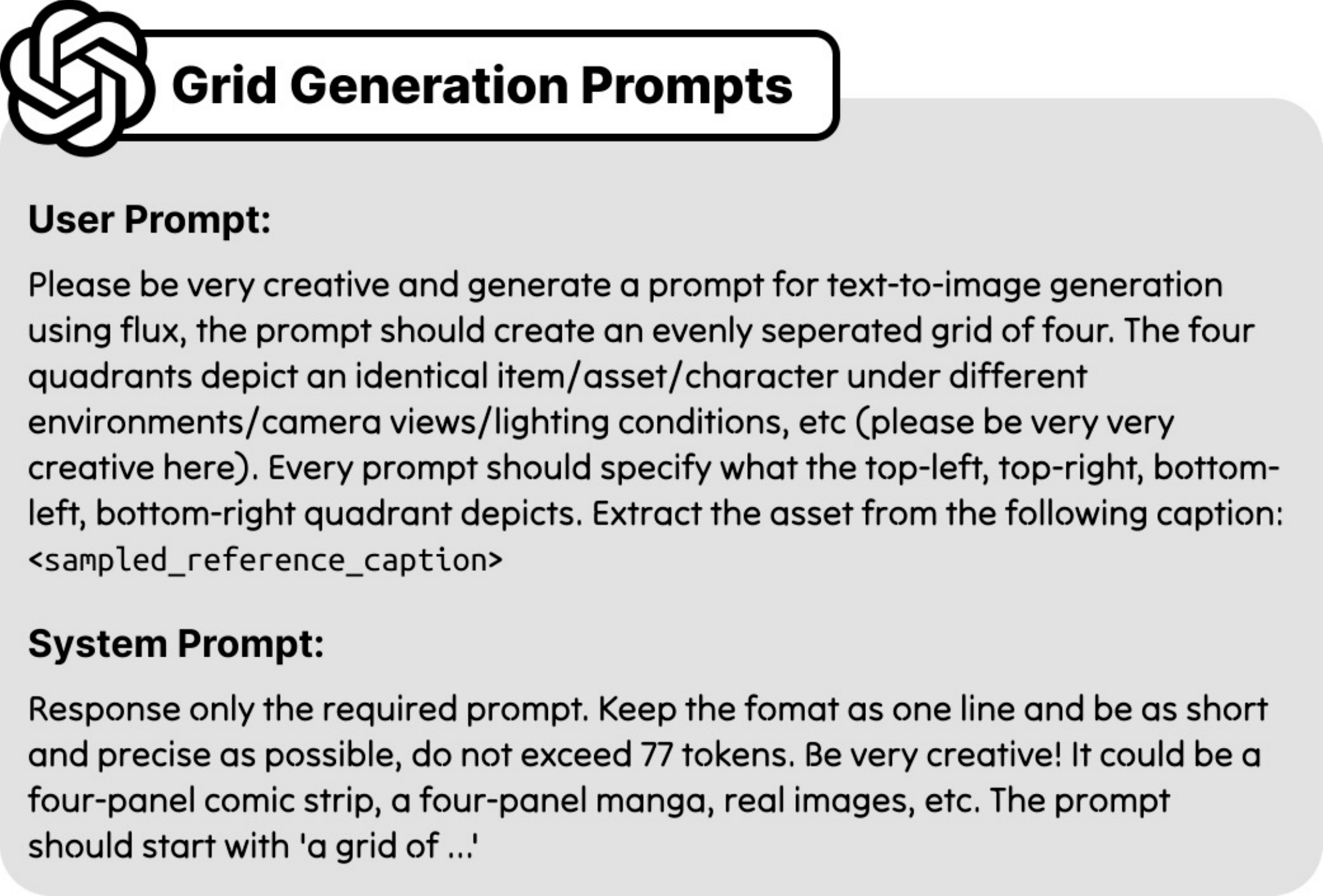}
}
\end{center}
\label{fig:grid_generation_prompts}
\end{figure}
\subsection{Data Curation Prompts}
\label{sec:data_curation_prompts}
For data curation, we employ Gemini-1.5.
To guide the vision-language model~(VLM) in focusing on identity preservation, we utilize Chain-of-Thought~(CoT) prompting~\cite{wei2022chainofthought}.
Specifically, we first instruct the VLM to identify the common object or character present in both images.
Next, we prompt it to describe each one in detail.
Finally, we ask the VLM to analyze whether they are identical and to provide a conclusive response.
We find that this CoT prompting significantly enhances the model's ability to concentrate on the identity and intricate details of the target object or character.
\begin{figure}[H]
\small
\begin{center}
{\small
\includegraphics[width=0.99\columnwidth]{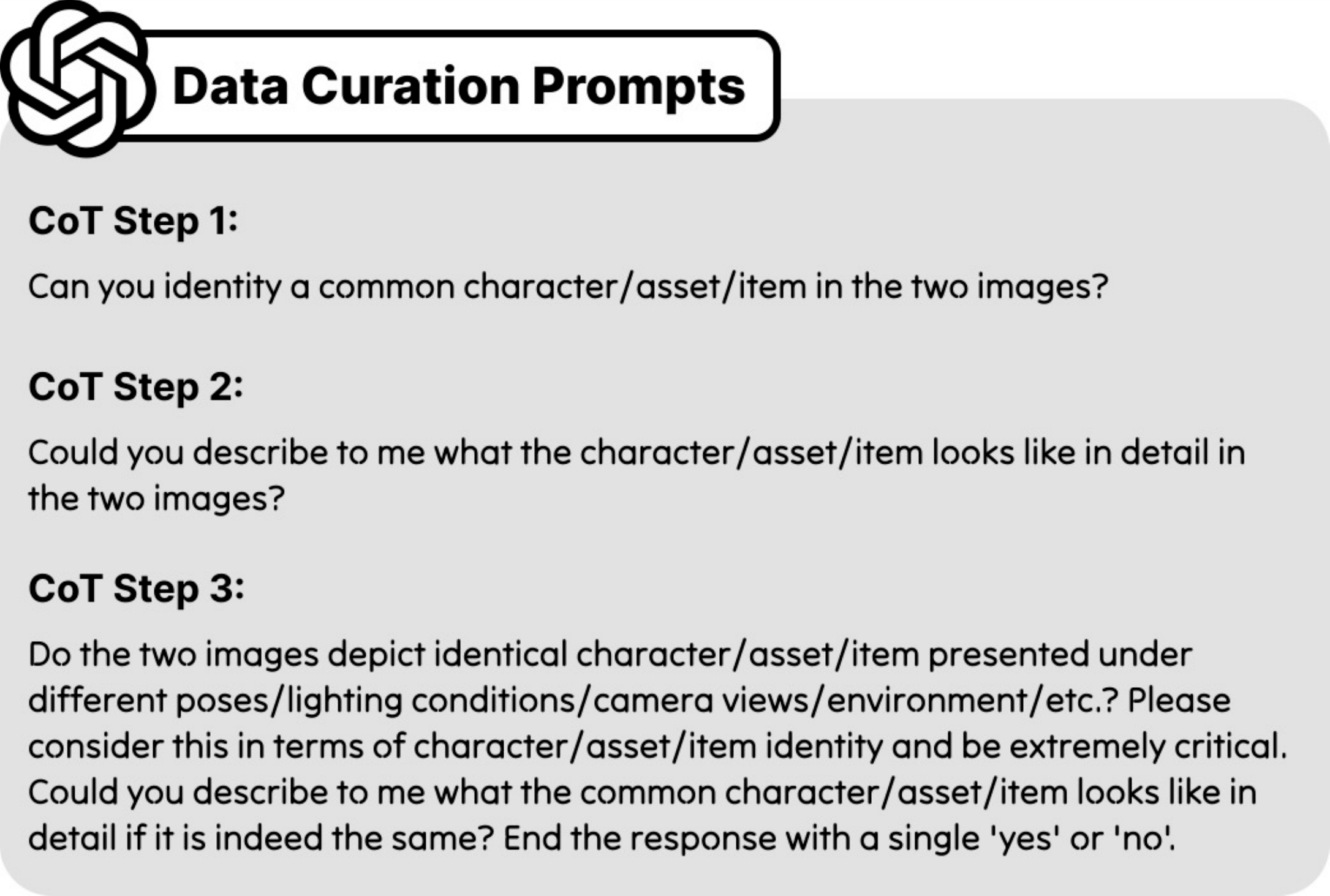}
}
\end{center}
\label{fig:data_curation_prompts}
\end{figure}
\subsection{Image Caption Prompts}
\label{sec:image_caption_prompts}
We provide two methods for prompting our model: using the description of the expected output~(Target Description) or InstructPix2Pix~\cite{brooks2022instructpix2pix}-type instructions~(Instruction).
\begin{figure}[H]
\small
\begin{center}
{\small
\includegraphics[width=0.99\columnwidth]{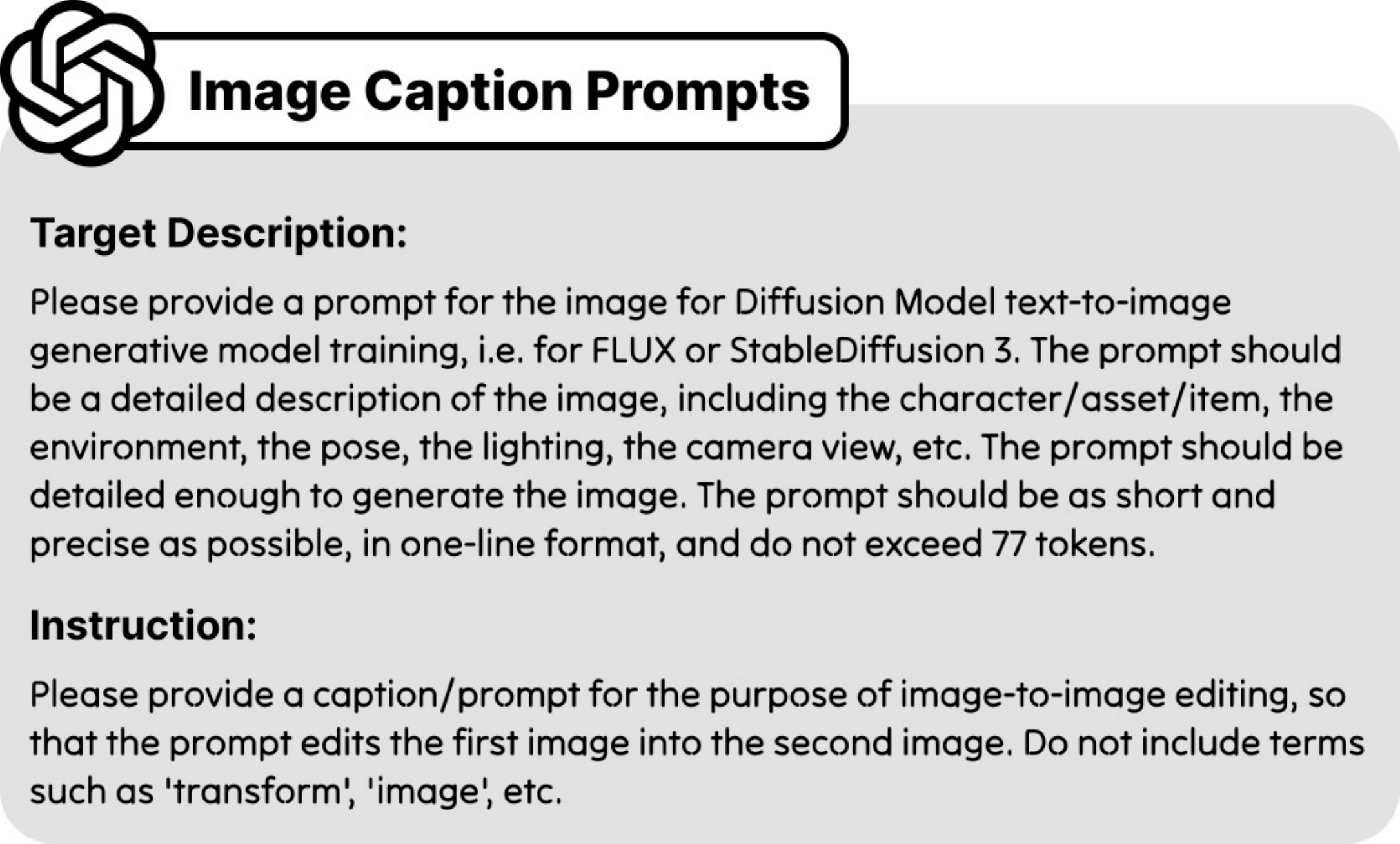}
}
\end{center}
\label{fig:image_caption_prompts}
\end{figure}
\section{GPT Evaluation Prompts}
\label{sec:gpt_evaluation_prompts}
\input{fig/gpt_evaluation_prompts}
We closely follow DreamBench++~\cite{peng2024dreambenchplus} in terms of our GPT evaluation.
In Fig.~\ref{fig:gpt_evaluation_prompts}, we demonstrate the prompts we use for evaluation, including our ``de-biased'' evaluation that penalizes ``copy-pasting'' effect.
\section{Additional Results}
\label{sec:additional_result}
\subsection{Additional Qualitative Comparisons}
\label{sec:more_qualitative_comparison}
In Fig.~\ref{fig:additional_qualitative_comparison}, we demonstrate more of the qualitative evaluation cases from the DreamBench++~\cite{peng2024dreambenchplus} benchmark.
\subsection{Additional Qualitative Results}
\label{sec:more_qualitative_result}
Due to space constraints in the main paper, we presented shortened prompts.
Here, we provide additional qualitative results in Fig.~\ref{fig:character_1}, Fig.~\ref{fig:character_2}, Fig.~\ref{fig:item_1}, Fig.~\ref{fig:item_2}, Fig.~\ref{fig:instruction_1} and Fig.~\ref{fig:relighting_1}, including the full prompts used for their generation.
These detailed captions capture various aspects of the images and offer deeper insights into how our model operates.
\subsection{Story Telling}
\label{sec:story_telling}
Our model exhibits the capability to generate simple comics and manga narratives, as demonstrated in Fig.~\ref{fig:comic_1} and Fig.~\ref{fig:comic_2}, where the conditioning image acts as the first panel. To create these storytelling sequences, we input the initial panel into GPT-4o, which generates a series of prompts centered around the main character from the input image.
These prompts are crafted to form a coherent story spanning 8--10 panels, with each prompt being contextually meaningful on its own.
Utilizing these prompts alongside the conditioning image, we generate the subsequent panels and finally align them to reconstruct a cohesive narrative.
\begin{figure}[H]
\small
\begin{center}
{\small
\includegraphics[width=0.99\columnwidth]{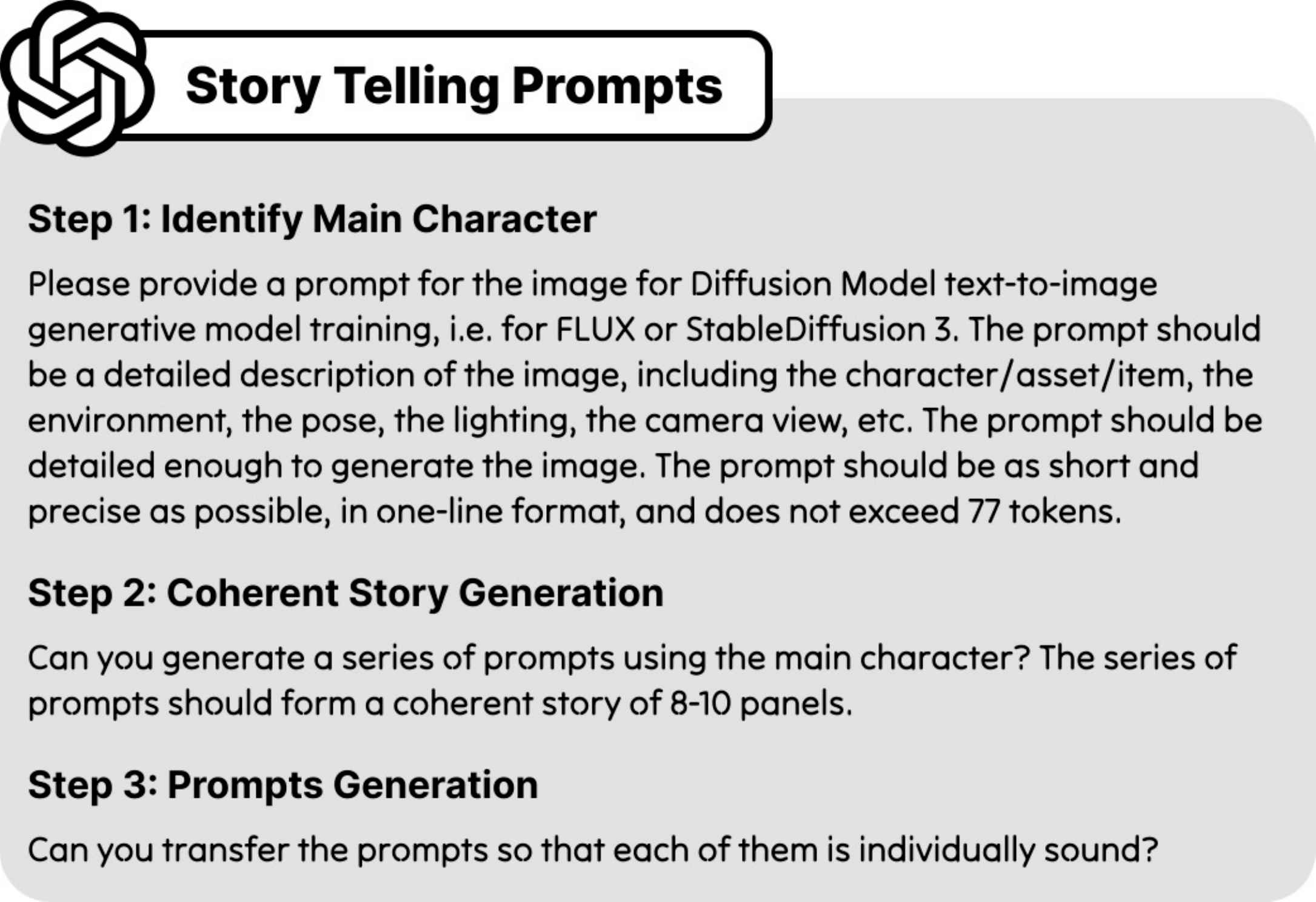}
}
\end{center}
\label{fig:story_telling_prompts}
\end{figure}

\section{Discussion on Scalability}
We acknowledge that the scalability of \acro{} is not fully explored within the scope of this paper.
However, we posit that \acro{} is inherently scalable along three key dimensions.
First, \acro{} can scale with advancements in the teacher model's grid generation capabilities and its in-context understanding of identity preservation.
Second, the scalability extends to the range of tasks we leverage; while this paper focuses on general adaptation tasks, a broader spectrum of applications remains open for exploration.
Third, \acro{} scales with the extent to which we harness foundation models.
Increased diversity and more meticulously curated data contribute to improved generalization of our model.
As foundation models—including base text-to-image generation models, language models~(LLMs), and vision-language models~(VLMs)—continue to evolve, \acro{} naturally benefits from these advancements without necessitating any modifications to the existing workflow.
A direct next step involves scaling the method to incorporate a significantly larger dataset and integrating forthcoming, more advanced foundation models.

\input{fig/additional_qualitative_comparison}
\input{fig/character_1}
\input{fig/character_2}
\input{fig/item_1}
\input{fig/item_2}
\input{fig/instruction_1}
\input{fig/relighting_1}
\input{fig/comic_1}
\input{fig/comic_2}

%% file: fig/gpt_evaluation_prompts.tex
\begin{figure*}[t]
\begin{center}
\centering
\includegraphics[width=0.99\linewidth]{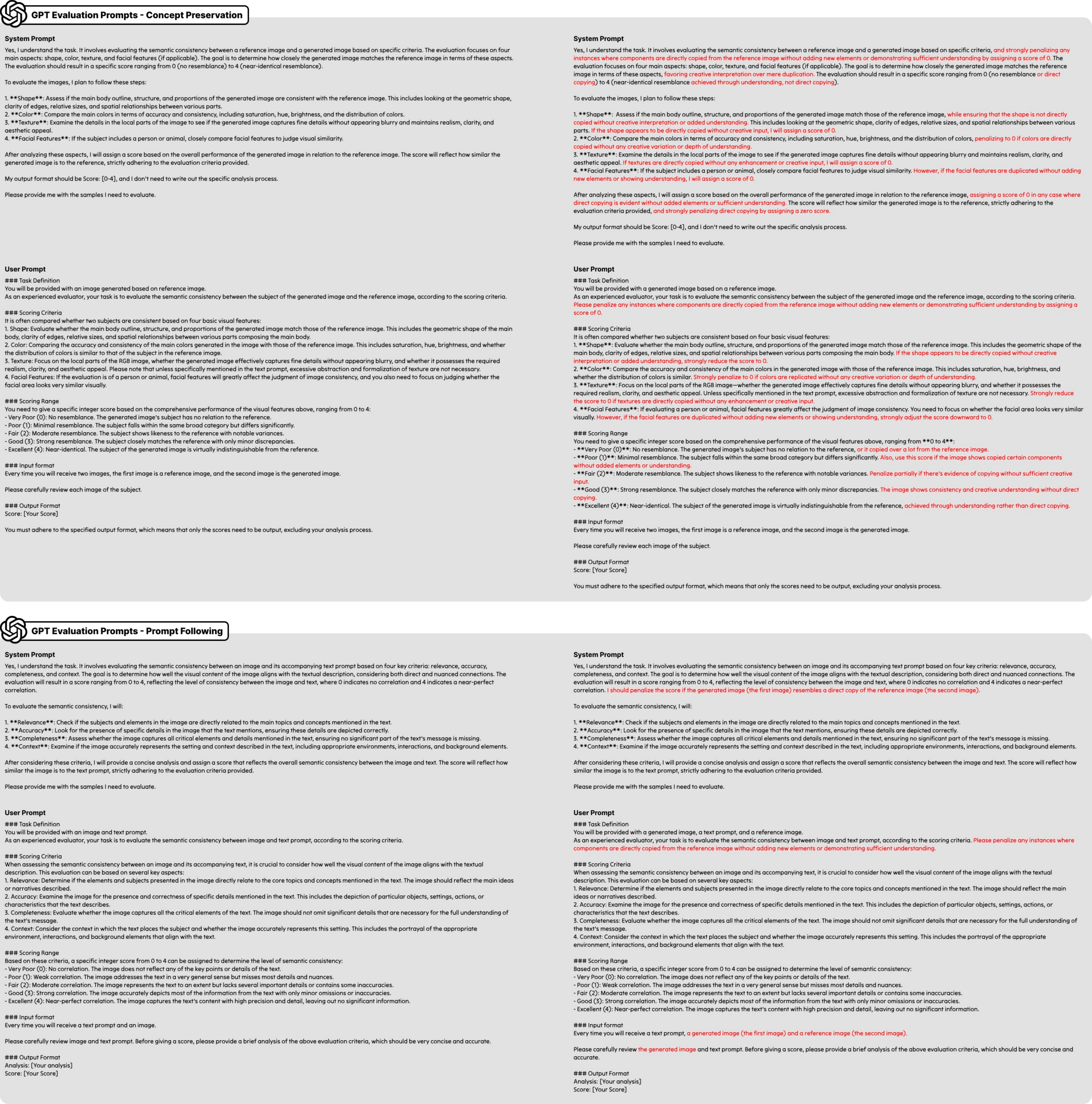}
\end{center}
\caption{
\textbf{GPT evaluation prompts } used across our evaluation, where the left shows the vanilla prompts from DreamBench++~\cite{peng2024dreambenchplus} and the right shows our modified ``de-biased'' prompts, which strongly penalizes ``copy-pasting'' effects without sufficient creative inputs.
We highlight our modified sentences in red.
}
\label{fig:gpt_evaluation_prompts}
\end{figure*}

%% file: fig/additional_qualitative_comparison.tex
\begin{figure*}[t]
\begin{center}
\centering
\includegraphics[width=0.99\linewidth]{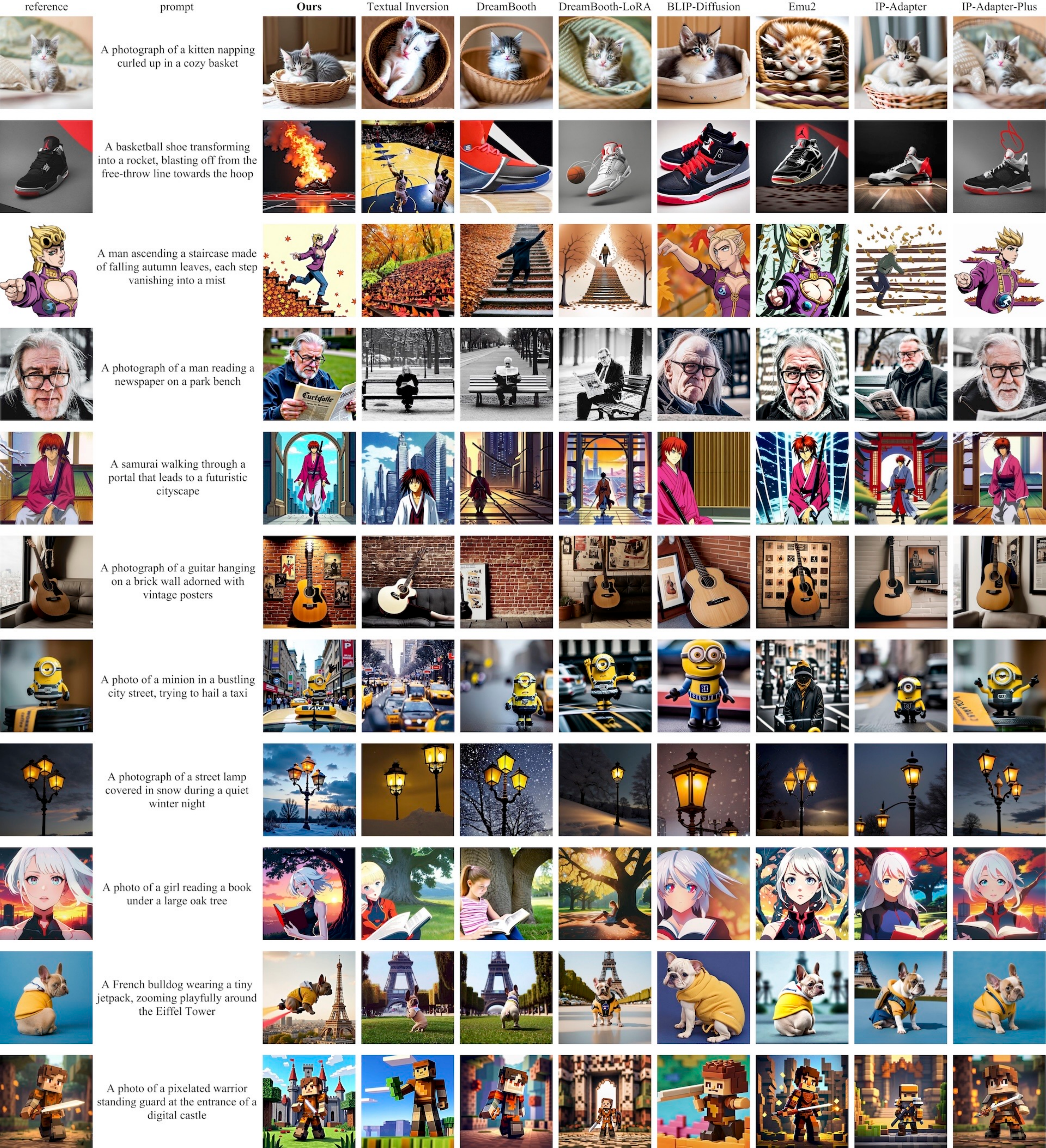}
\end{center}
\caption{
\textbf{Additional qualitative comparison. }
}
\label{fig:additional_qualitative_comparison}
\end{figure*}

%% file: fig/character_1.tex
\begin{figure*}[t]
\begin{center}
\centering
\includegraphics[width=0.99\linewidth]{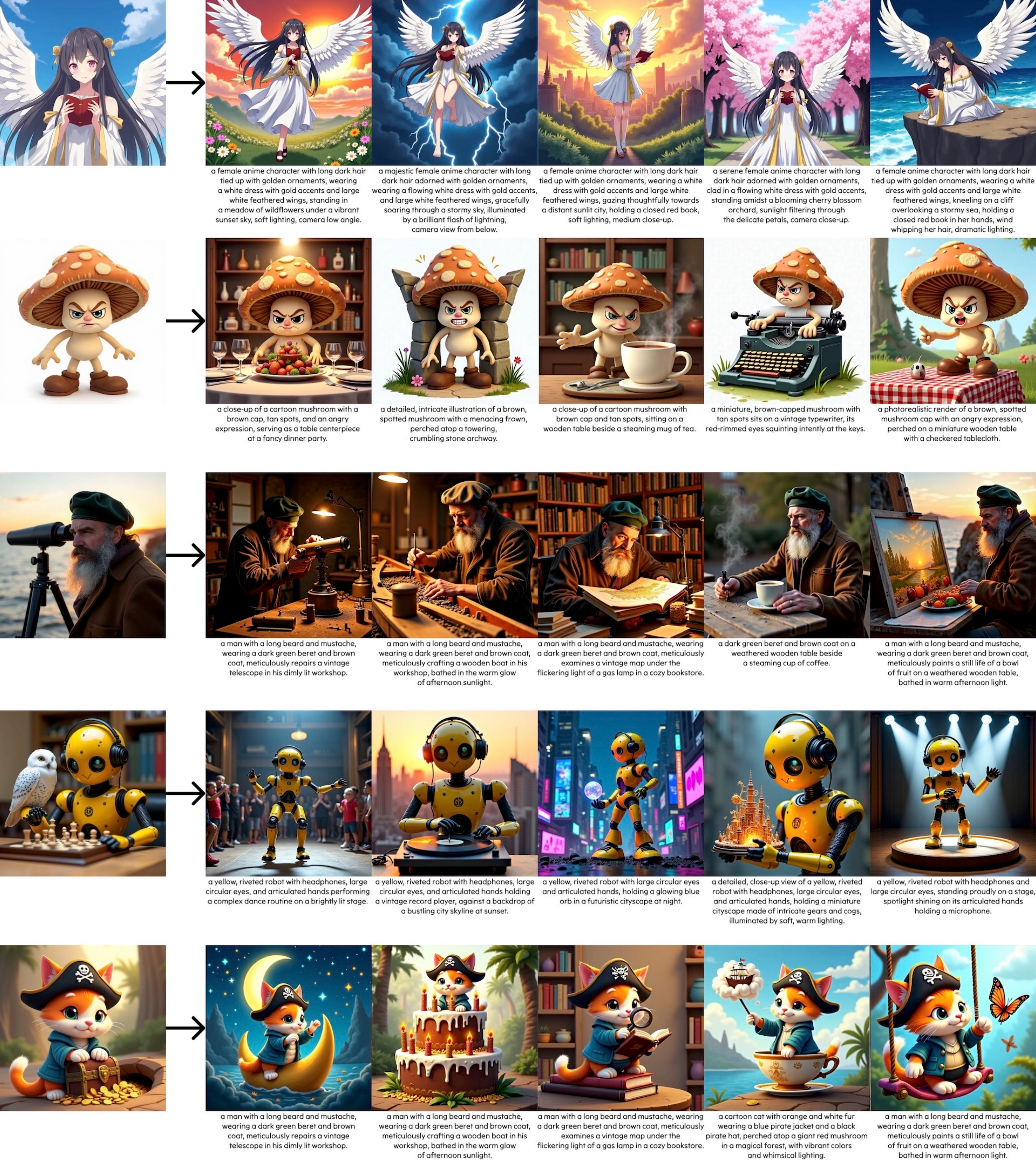}
\end{center}
\caption{
\textbf{Additional character identity preserving results. }
}
\label{fig:character_1}
\end{figure*}

%% file: fig/character_2.tex
\begin{figure*}[t]
\begin{center}
\centering
\includegraphics[width=0.99\linewidth]{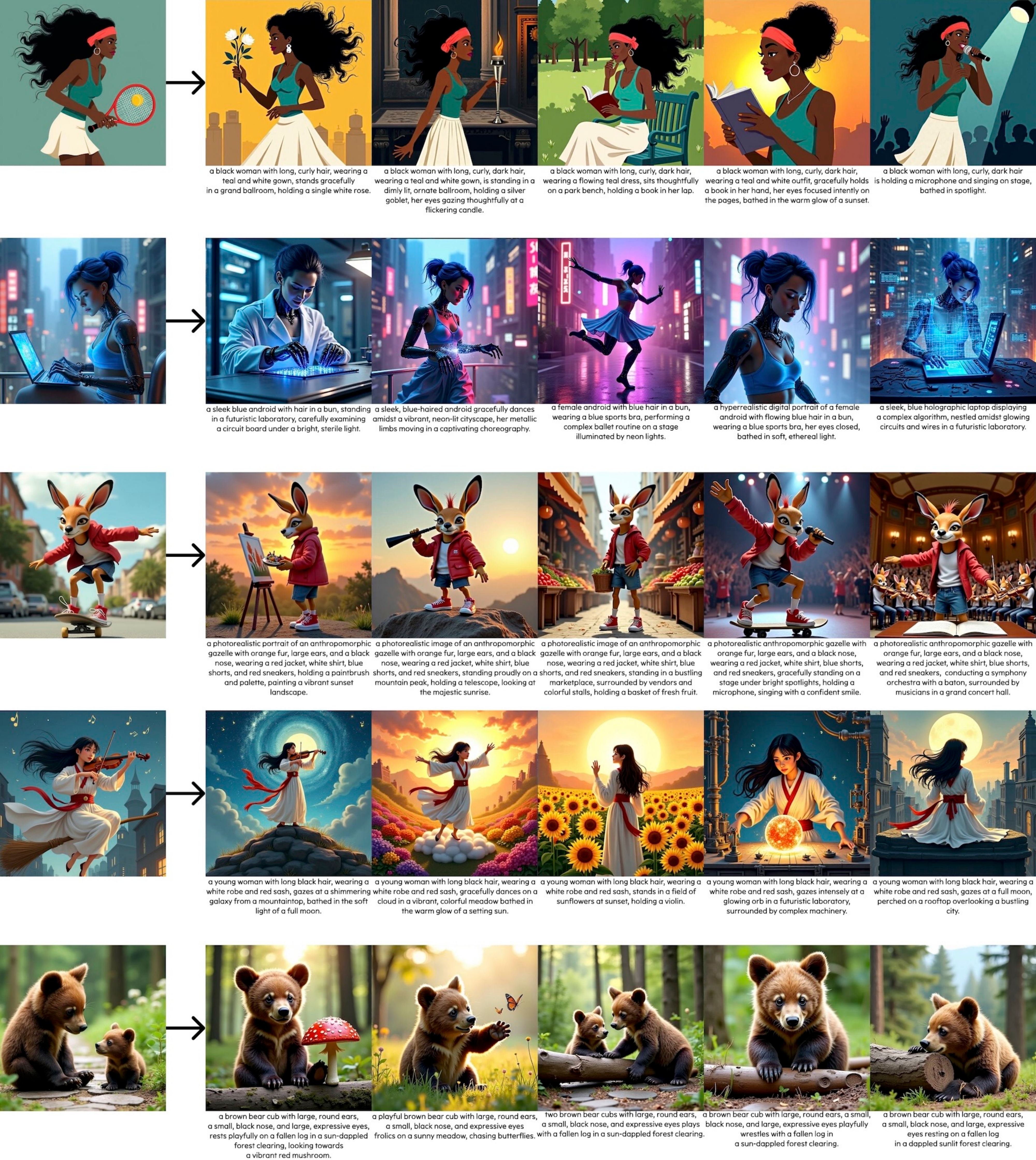}
\end{center}
\caption{
\textbf{Additional character identity preserving results. }
}
\label{fig:character_2}
\end{figure*}

%% file: fig/item_1.tex
\begin{figure*}[t]
\begin{center}
\centering
\includegraphics[width=0.99\linewidth]{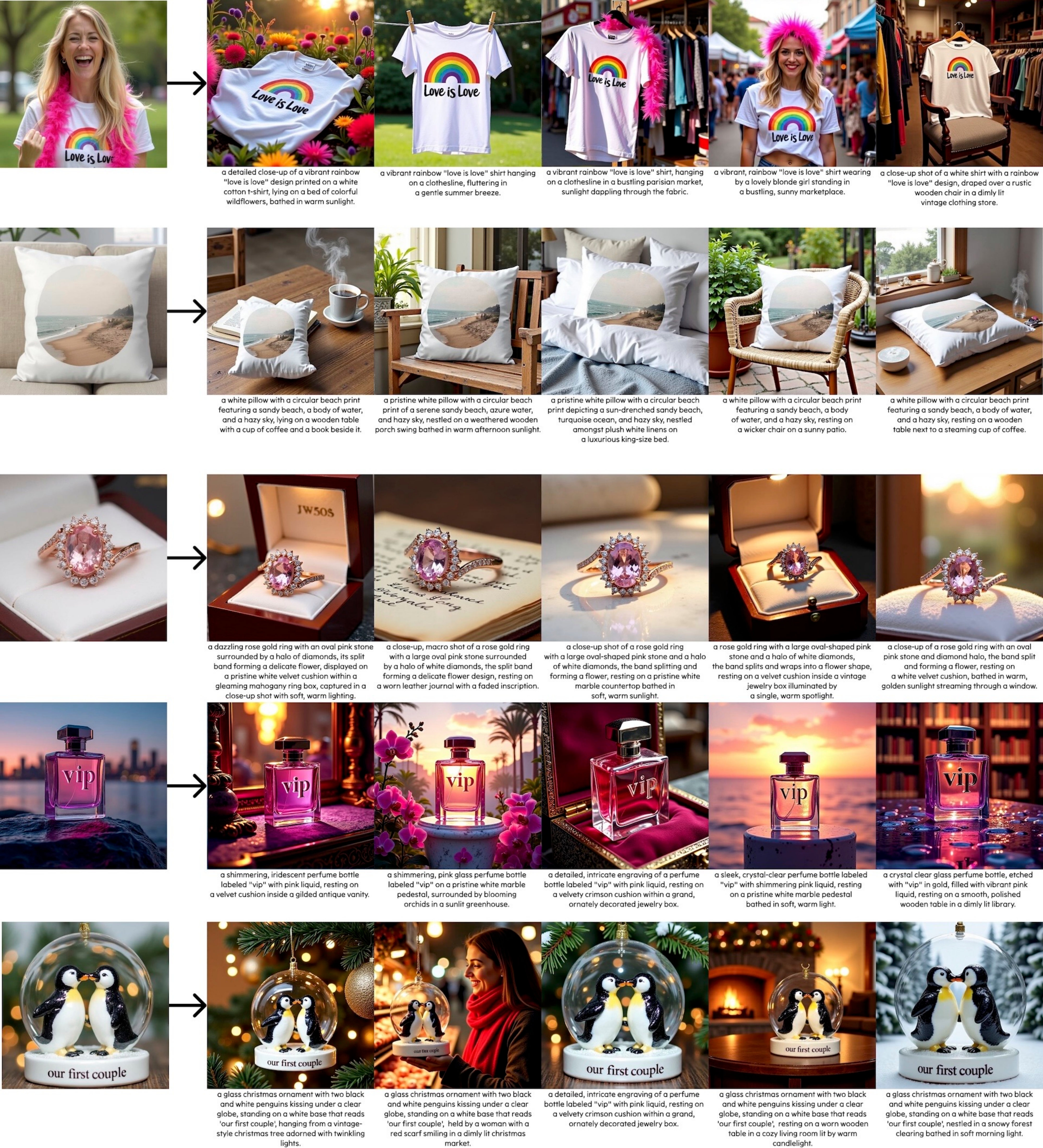}
\end{center}
\caption{
\textbf{Additional object/item identity preserving results. }
}
\label{fig:item_1}
\end{figure*}

%% file: fig/item_2.tex
\begin{figure*}[t]
\begin{center}
\centering
\includegraphics[width=0.99\linewidth]{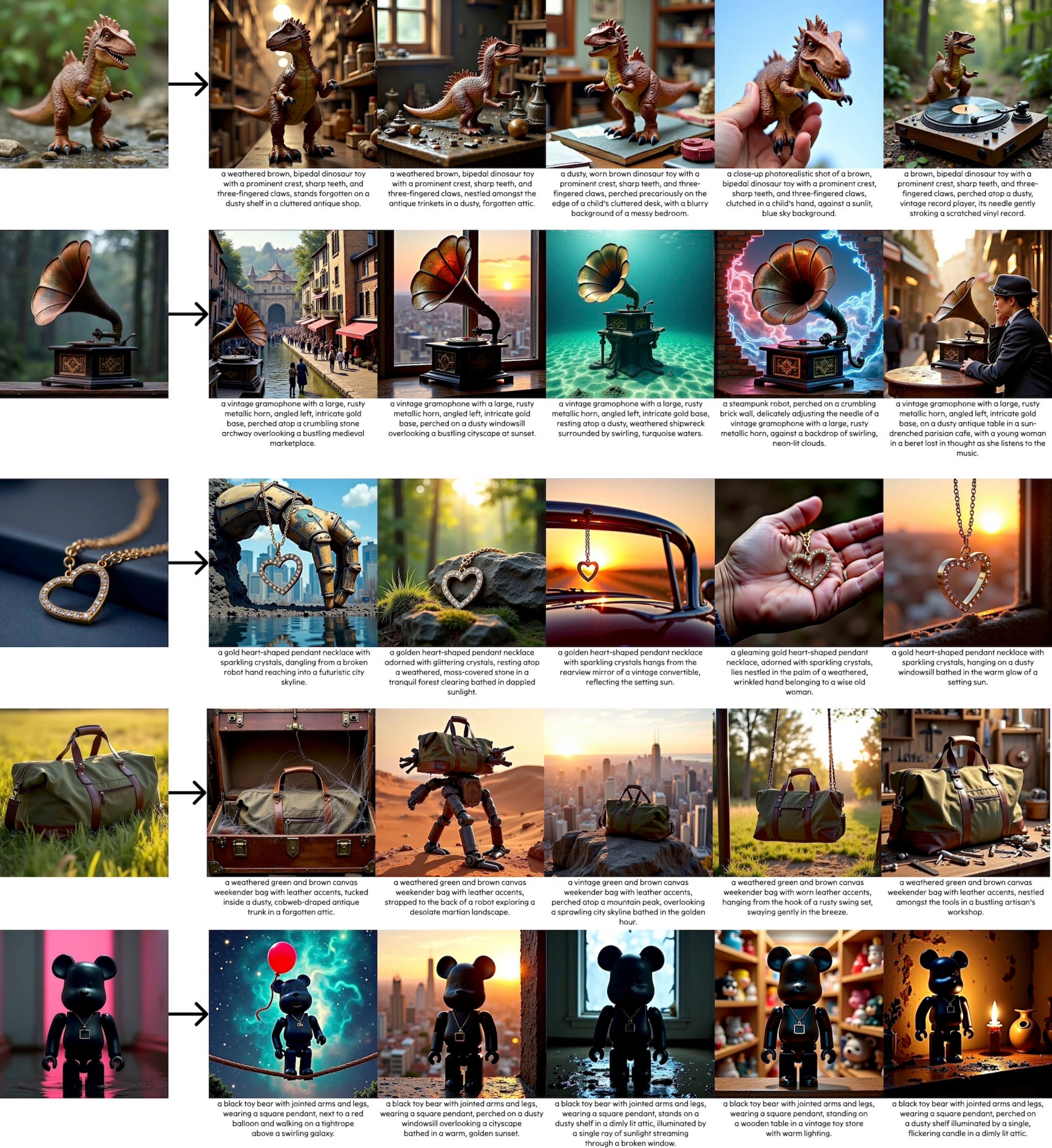}
\end{center}
\caption{
\textbf{Additional object/item identity preserving results. }
}
\label{fig:item_2}
\end{figure*}

%% file: fig/instruction_1.tex
\begin{figure*}[t]
\begin{center}
\centering
\includegraphics[width=0.99\linewidth]{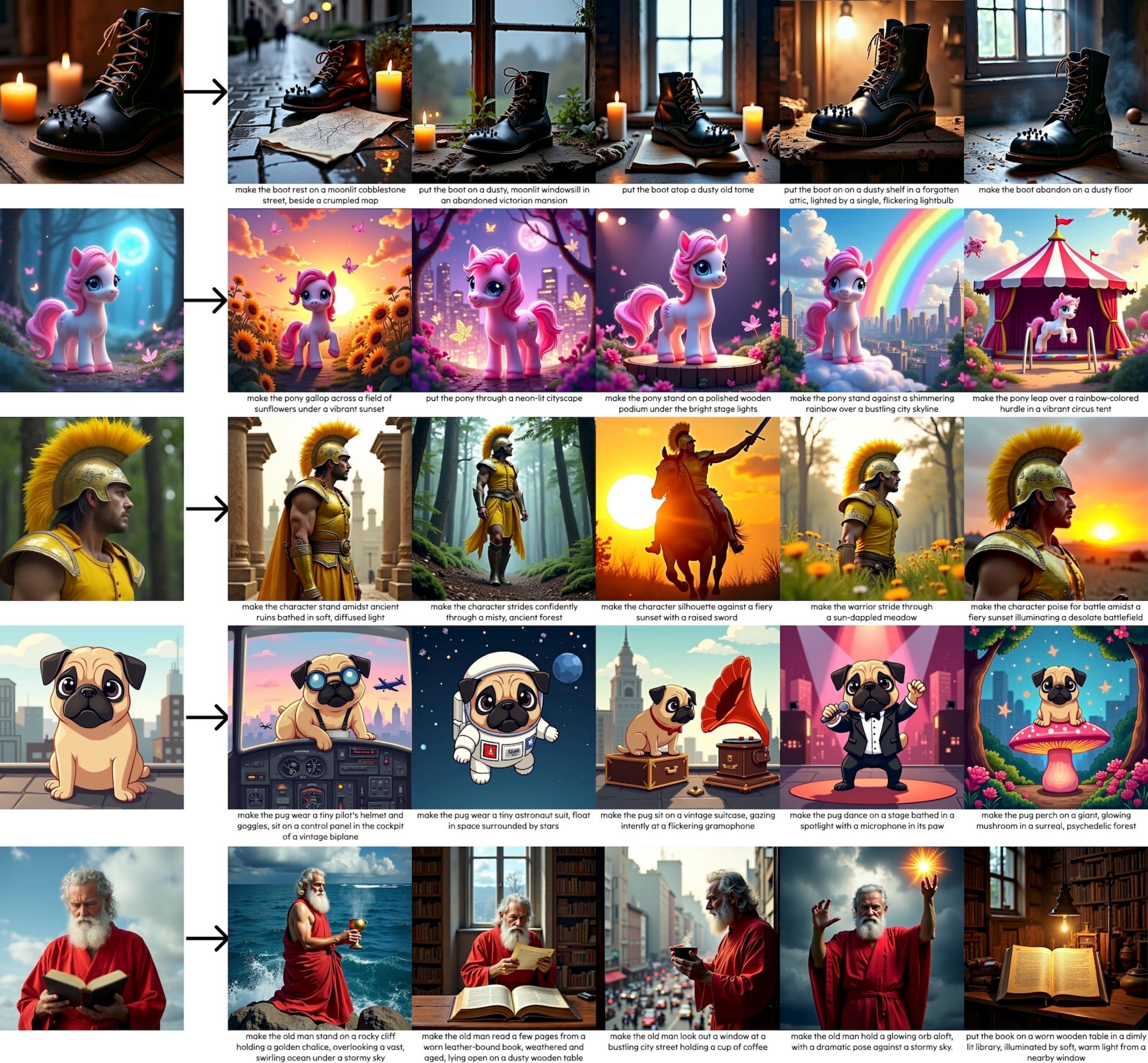}
\end{center}
\caption{
\textbf{Additional instruction prompting results. }
}
\label{fig:instruction_1}
\end{figure*}

%% file: fig/relighting_1.tex
\begin{figure*}[t]
\begin{center}
\centering
\includegraphics[width=0.99\linewidth]{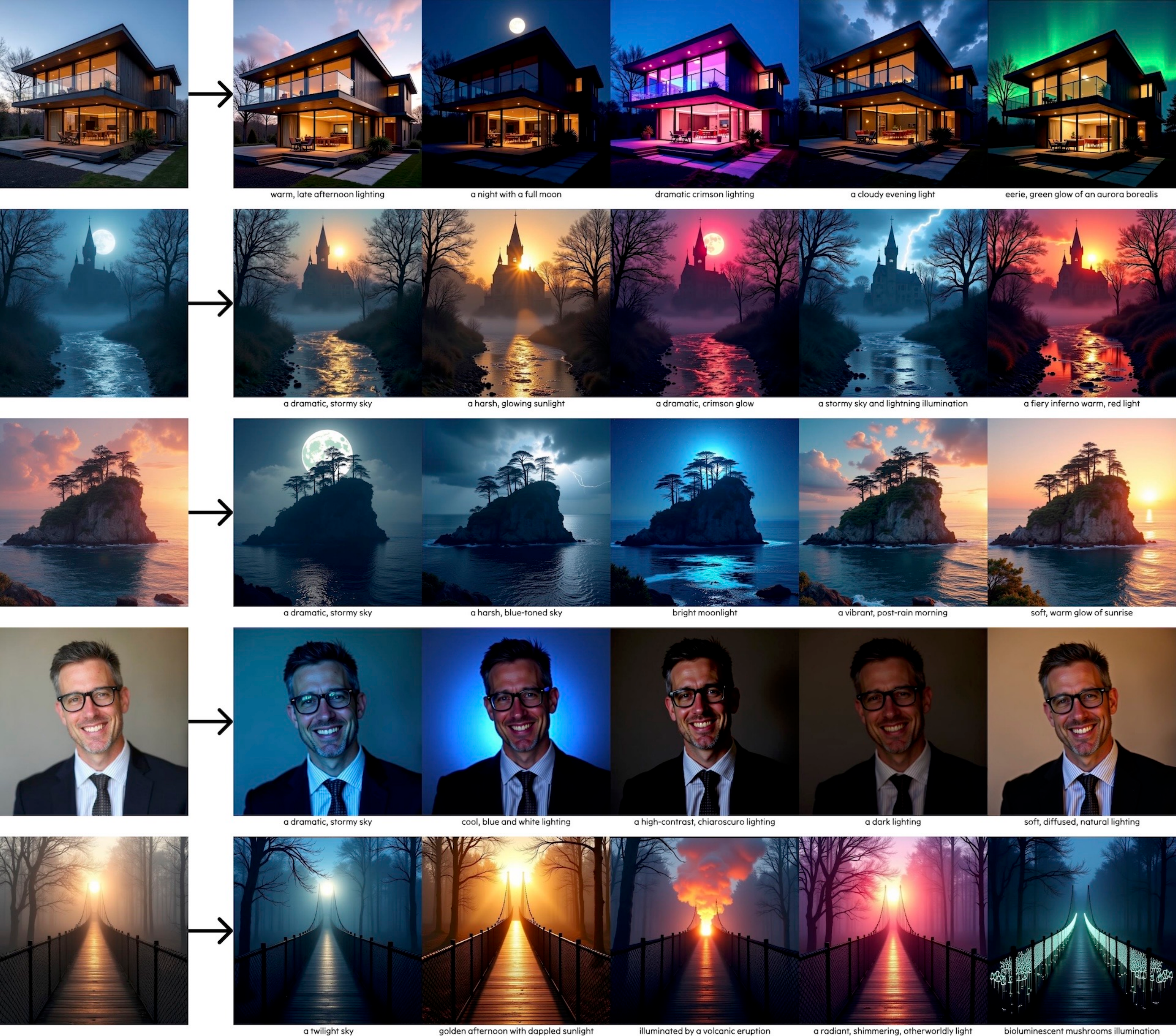}
\end{center}
\caption{
\textbf{Additional relighting results. }
}
\label{fig:relighting_1}
\end{figure*}

%% file: fig/comic_1.tex
\begin{figure*}[t]
\begin{center}
\centering
\includegraphics[width=0.99\linewidth]{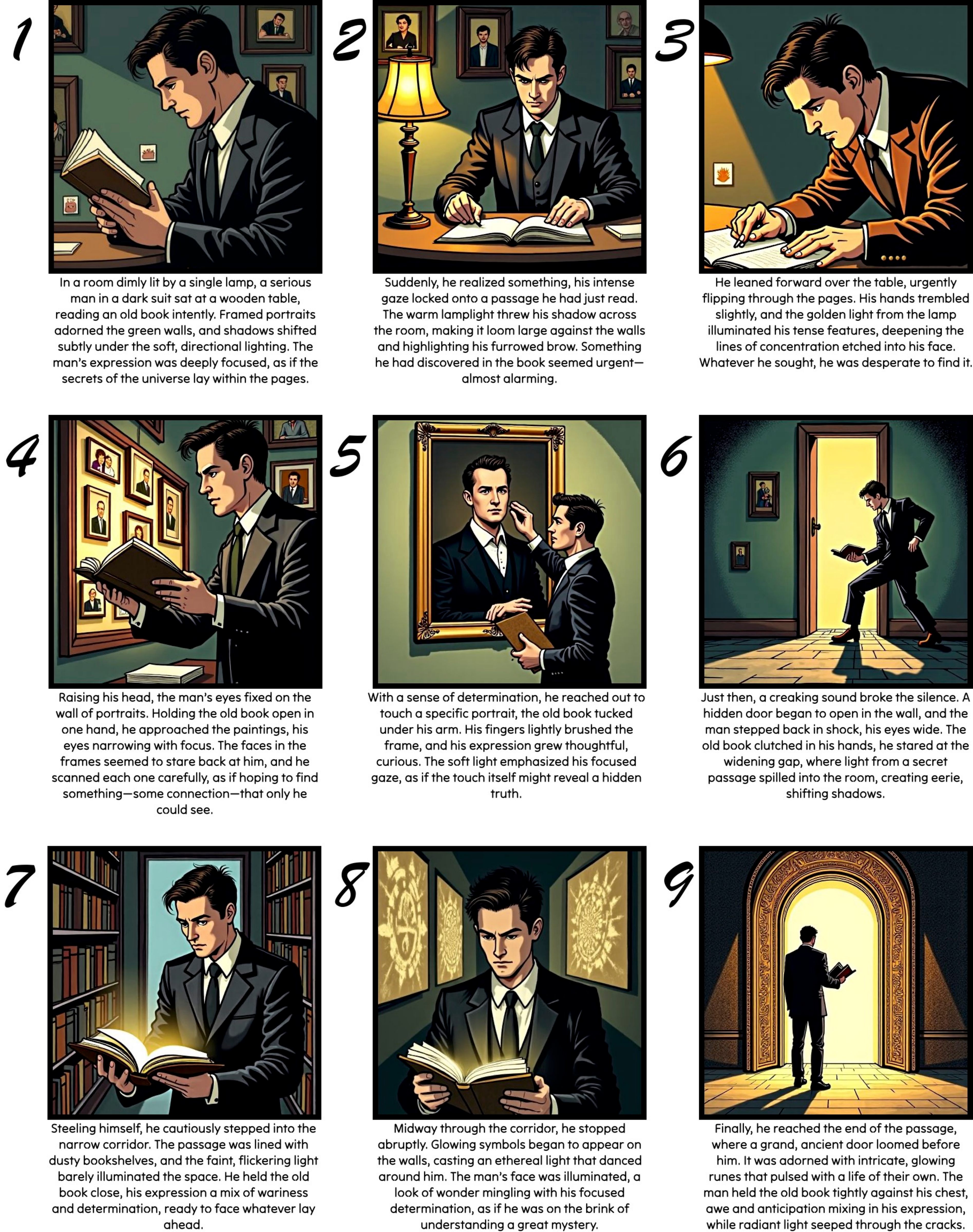}
\end{center}
\caption{
\textbf{Comic generation example 1. }
The conditioned image is the first panel.
}
\label{fig:comic_1}
\end{figure*}

%% file: fig/comic_2.tex
\begin{figure*}[t]
\begin{center}
\centering
\includegraphics[width=0.99\linewidth]{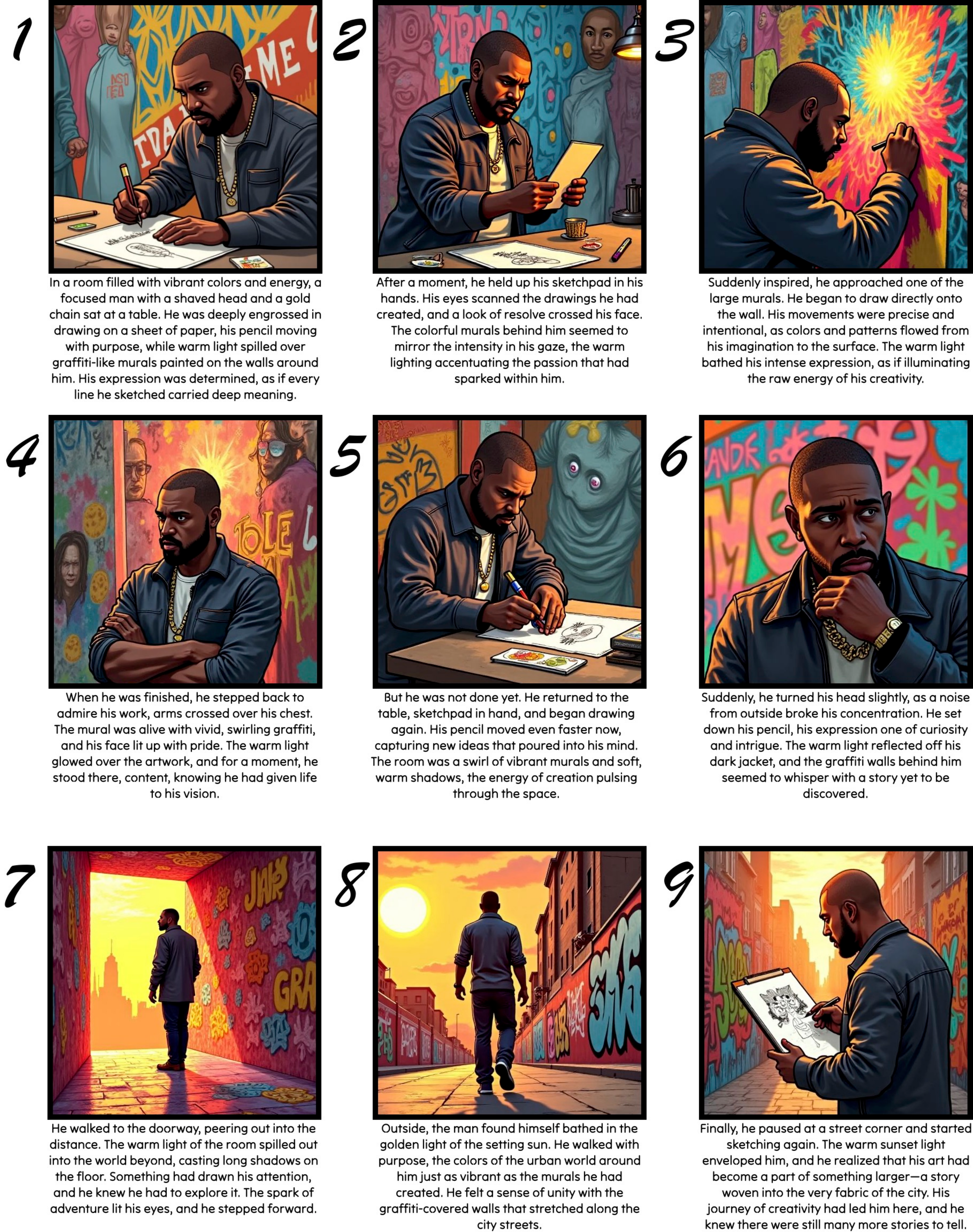}
\end{center}
\caption{
\textbf{Comic generation example 2. }
The conditioned image is the first panel.
}
\label{fig:comic_2}
\end{figure*}